\documentclass[letterpaper, 10 pt, conference]{ieeeconf}  

\IEEEoverridecommandlockouts                              

\usepackage[T1]{fontenc}    
\usepackage{url}            
\usepackage{booktabs}       
\usepackage{amsfonts}       
\usepackage{nicefrac}       
\usepackage{microtype}      
\usepackage{xcolor}         
\usepackage{amsmath}
\usepackage{xfp}            
\usepackage{graphicx}    
\usepackage{caption}
\usepackage{subcaption}
\usepackage{multirow}
\usepackage{wrapfig}
\usepackage{makecell}
\usepackage[utf8]{inputenc}
\usepackage{amssymb}
\usepackage{dblfloatfix}
\usepackage{placeins}
\usepackage[noadjust]{cite}  
\makeatletter
\let\NAT@parse\undefined
\makeatother
\usepackage[hidelinks,breaklinks]{hyperref}

\makeatletter
\newcommand{\resdef}[4]{\expandafter\gdef\csname res@#1@#2@#3\endcsname{#4}}
\newcommand{\res}[3]{%
  \ifcsname res@#1@#2@#3\endcsname
    \csname res@#1@#2@#3\endcsname
  \else
    \textcolor{red}{\textbf{??}}%
    \PackageWarning{results}{MISSING RESULT: #1/#2/#3}%
  \fi}
\makeatother

\newcommand{\resfail}{---}

\resdef{rawcd}{fastlio2}{cloister}{41.45}
\resdef{rawcd}{fastlio2}{matheasy}{34.74}
\resdef{rawcd}{fastlio2}{mathhard}{37.70}
\resdef{rawcd}{fastlio2}{quadeasy}{27.85}
\resdef{rawcd}{fastlio2}{quadmid}{31.71}
\resdef{rawcd}{fastlio2}{quadhard}{37.74}
\resdef{rawcd}{fastlio2}{canteenday}{35.76}
\resdef{rawcd}{fastlio2}{canteennight}{35.03}
\resdef{rawcd}{fastlio2}{gardenday}{34.89}
\resdef{rawcd}{fastlio2}{gardennight}{30.84}

\resdef{rawcd}{hba}{cloister}{48.45}
\resdef{rawcd}{hba}{matheasy}{34.74}
\resdef{rawcd}{hba}{mathhard}{39.11}
\resdef{rawcd}{hba}{quadeasy}{29.93}
\resdef{rawcd}{hba}{quadmid}{38.54}
\resdef{rawcd}{hba}{quadhard}{42.22}
\resdef{rawcd}{hba}{canteenday}{35.67}
\resdef{rawcd}{hba}{canteennight}{35.03}
\resdef{rawcd}{hba}{gardenday}{36.11}
\resdef{rawcd}{hba}{gardennight}{31.61}

\resdef{rawcd}{balm}{cloister}{41.23}
\resdef{rawcd}{balm}{matheasy}{34.02}
\resdef{rawcd}{balm}{mathhard}{37.45}
\resdef{rawcd}{balm}{quadeasy}{24.04}
\resdef{rawcd}{balm}{quadmid}{27.00}
\resdef{rawcd}{balm}{quadhard}{39.05}
\resdef{rawcd}{balm}{canteenday}{34.58}
\resdef{rawcd}{balm}{canteennight}{33.41}
\resdef{rawcd}{balm}{gardenday}{30.70}
\resdef{rawcd}{balm}{gardennight}{27.23}

\resdef{rawcd}{gtpose}{cloister}{46.63}
\resdef{rawcd}{gtpose}{matheasy}{33.34}
\resdef{rawcd}{gtpose}{mathhard}{42.44}
\resdef{rawcd}{gtpose}{quadeasy}{29.72}
\resdef{rawcd}{gtpose}{quadmid}{34.11}
\resdef{rawcd}{gtpose}{quadhard}{40.76}
\resdef{rawcd}{gtpose}{canteenday}{36.29}
\resdef{rawcd}{gtpose}{canteennight}{34.67}
\resdef{rawcd}{gtpose}{gardenday}{31.63}
\resdef{rawcd}{gtpose}{gardennight}{34.58}

\resdef{rawcd}{ablnohiernocube}{mathhard}{67.8}
\resdef{rawcd}{ablnohiernocube}{quadhard}{65.3}
\resdef{rawcd}{ablnohiernocube}{canteenday}{43.0}

\resdef{rawcd}{ablnohier}{mathhard}{48.9}
\resdef{rawcd}{ablnohier}{quadhard}{32.9}
\resdef{rawcd}{ablnohier}{canteenday}{37.9}

\resdef{rawcd}{ablnoterm}{mathhard}{44.6}
\resdef{rawcd}{ablnoterm}{quadhard}{46.6}
\resdef{rawcd}{ablnoterm}{canteenday}{43.8}

\resdef{rawcd}{ablnosurrogate}{mathhard}{54.8}
\resdef{rawcd}{ablnosurrogate}{quadhard}{40.1}
\resdef{rawcd}{ablnosurrogate}{canteenday}{59.7}

\resdef{rawcd}{ablnerfhier}{mathhard}{37.3}
\resdef{rawcd}{ablnerfhier}{quadhard}{31.7}
\resdef{rawcd}{ablnerfhier}{canteenday}{34.8}

\resdef{rawcd}{ablhash}{mathhard}{35.8}
\resdef{rawcd}{ablhash}{quadhard}{30.0}
\resdef{rawcd}{ablhash}{canteenday}{35.6}

\resdef{rawcd}{ablours}{mathhard}{34.9}
\resdef{rawcd}{ablours}{quadhard}{29.6}
\resdef{rawcd}{ablours}{canteenday}{34.5}

\resdef{rawcd}{ours}{cloister}{43.84}
\resdef{rawcd}{ours}{matheasy}{34.16}
\resdef{rawcd}{ours}{mathhard}{34.92}
\resdef{rawcd}{ours}{quadeasy}{23.80}
\resdef{rawcd}{ours}{quadmid}{26.19}
\resdef{rawcd}{ours}{quadhard}{29.59}
\resdef{rawcd}{ours}{canteenday}{34.51}
\resdef{rawcd}{ours}{canteennight}{33.12}
\resdef{rawcd}{ours}{gardenday}{30.44}
\resdef{rawcd}{ours}{gardennight}{26.92}

\resdef{rendcd}{pinslam}{cloister}{17.14}
\resdef{rendcd}{pinslam}{matheasy}{16.89}
\resdef{rendcd}{pinslam}{mathhard}{\resfail}
\resdef{rendcd}{pinslam}{quadeasy}{17.21}
\resdef{rendcd}{pinslam}{quadmid}{16.49}
\resdef{rendcd}{pinslam}{quadhard}{17.27}
\resdef{rendcd}{pinslam}{canteenday}{14.18}
\resdef{rendcd}{pinslam}{canteennight}{13.63}
\resdef{rendcd}{pinslam}{gardenday}{15.66}
\resdef{rendcd}{pinslam}{gardennight}{16.07}

\resdef{rendcd}{shine}{cloister}{16.31}
\resdef{rendcd}{shine}{matheasy}{16.28}
\resdef{rendcd}{shine}{mathhard}{15.88}
\resdef{rendcd}{shine}{quadeasy}{14.23}
\resdef{rendcd}{shine}{quadmid}{15.48}
\resdef{rendcd}{shine}{quadhard}{16.30}
\resdef{rendcd}{shine}{canteenday}{13.21}
\resdef{rendcd}{shine}{canteennight}{12.99}
\resdef{rendcd}{shine}{gardenday}{14.96}
\resdef{rendcd}{shine}{gardennight}{14.83}

\resdef{rendcd}{fourdndf}{cloister}{15.42}
\resdef{rendcd}{fourdndf}{matheasy}{15.66}
\resdef{rendcd}{fourdndf}{mathhard}{\resfail}
\resdef{rendcd}{fourdndf}{quadeasy}{16.11}
\resdef{rendcd}{fourdndf}{quadmid}{15.48}
\resdef{rendcd}{fourdndf}{quadhard}{15.68}
\resdef{rendcd}{fourdndf}{canteenday}{12.38}
\resdef{rendcd}{fourdndf}{canteennight}{12.13}
\resdef{rendcd}{fourdndf}{gardenday}{13.89}
\resdef{rendcd}{fourdndf}{gardennight}{14.96}

\resdef{rendcd}{ours}{cloister}{15.56}
\resdef{rendcd}{ours}{matheasy}{15.22}
\resdef{rendcd}{ours}{mathhard}{15.88}
\resdef{rendcd}{ours}{quadeasy}{11.97}
\resdef{rendcd}{ours}{quadmid}{12.80}
\resdef{rendcd}{ours}{quadhard}{15.47}
\resdef{rendcd}{ours}{canteenday}{12.70}
\resdef{rendcd}{ours}{canteennight}{12.36}
\resdef{rendcd}{ours}{gardenday}{12.06}
\resdef{rendcd}{ours}{gardennight}{12.36}

\resdef{rendf1}{pinslam}{cloister}{56.1}
\resdef{rendf1}{pinslam}{matheasy}{50.0}
\resdef{rendf1}{pinslam}{mathhard}{\resfail}
\resdef{rendf1}{pinslam}{quadeasy}{63.6}
\resdef{rendf1}{pinslam}{quadmid}{41.3}
\resdef{rendf1}{pinslam}{quadhard}{47.5}
\resdef{rendf1}{pinslam}{canteenday}{34.5}
\resdef{rendf1}{pinslam}{canteennight}{32.4}
\resdef{rendf1}{pinslam}{gardenday}{55.6}
\resdef{rendf1}{pinslam}{gardennight}{58.6}

\resdef{rendf1}{shine}{cloister}{65.2}
\resdef{rendf1}{shine}{matheasy}{71.8}
\resdef{rendf1}{shine}{mathhard}{59.0}
\resdef{rendf1}{shine}{quadeasy}{82.7}
\resdef{rendf1}{shine}{quadmid}{73.2}
\resdef{rendf1}{shine}{quadhard}{62.2}
\resdef{rendf1}{shine}{canteenday}{52.6}
\resdef{rendf1}{shine}{canteennight}{50.1}
\resdef{rendf1}{shine}{gardenday}{65.3}
\resdef{rendf1}{shine}{gardennight}{69.9}

\resdef{rendf1}{fourdndf}{cloister}{51.9}
\resdef{rendf1}{fourdndf}{matheasy}{44.5}
\resdef{rendf1}{fourdndf}{mathhard}{\resfail}
\resdef{rendf1}{fourdndf}{quadeasy}{55.9}
\resdef{rendf1}{fourdndf}{quadmid}{49.2}
\resdef{rendf1}{fourdndf}{quadhard}{45.2}
\resdef{rendf1}{fourdndf}{canteenday}{35.1}
\resdef{rendf1}{fourdndf}{canteennight}{34.5}
\resdef{rendf1}{fourdndf}{gardenday}{50.4}
\resdef{rendf1}{fourdndf}{gardennight}{51.5}

\resdef{rendf1}{ours}{cloister}{59.0}
\resdef{rendf1}{ours}{matheasy}{79.7}
\resdef{rendf1}{ours}{mathhard}{77.7}
\resdef{rendf1}{ours}{quadeasy}{89.6}
\resdef{rendf1}{ours}{quadmid}{87.8}
\resdef{rendf1}{ours}{quadhard}{82.5}
\resdef{rendf1}{ours}{canteenday}{61.4}
\resdef{rendf1}{ours}{canteennight}{60.0}
\resdef{rendf1}{ours}{gardenday}{69.3}
\resdef{rendf1}{ours}{gardennight}{73.8}

\resdef{rendrmse}{pinslam}{cloister}{7.73}
\resdef{rendrmse}{pinslam}{matheasy}{6.32}
\resdef{rendrmse}{pinslam}{mathhard}{\resfail}
\resdef{rendrmse}{pinslam}{quadeasy}{6.53}
\resdef{rendrmse}{pinslam}{quadmid}{5.83}
\resdef{rendrmse}{pinslam}{quadhard}{6.39}
\resdef{rendrmse}{pinslam}{canteenday}{5.34}
\resdef{rendrmse}{pinslam}{canteennight}{5.17}
\resdef{rendrmse}{pinslam}{gardenday}{5.99}
\resdef{rendrmse}{pinslam}{gardennight}{6.07}

\resdef{rendrmse}{shine}{cloister}{8.27}
\resdef{rendrmse}{shine}{matheasy}{7.86}
\resdef{rendrmse}{shine}{mathhard}{7.16}
\resdef{rendrmse}{shine}{quadeasy}{6.73}
\resdef{rendrmse}{shine}{quadmid}{7.12}
\resdef{rendrmse}{shine}{quadhard}{7.38}
\resdef{rendrmse}{shine}{canteenday}{6.38}
\resdef{rendrmse}{shine}{canteennight}{6.30}
\resdef{rendrmse}{shine}{gardenday}{7.07}
\resdef{rendrmse}{shine}{gardennight}{6.97}

\resdef{rendrmse}{fourdndf}{cloister}{7.17}
\resdef{rendrmse}{fourdndf}{matheasy}{6.28}
\resdef{rendrmse}{fourdndf}{mathhard}{\resfail}
\resdef{rendrmse}{fourdndf}{quadeasy}{6.73}
\resdef{rendrmse}{fourdndf}{quadmid}{6.42}
\resdef{rendrmse}{fourdndf}{quadhard}{6.34}
\resdef{rendrmse}{fourdndf}{canteenday}{5.08}
\resdef{rendrmse}{fourdndf}{canteennight}{5.06}
\resdef{rendrmse}{fourdndf}{gardenday}{5.90}
\resdef{rendrmse}{fourdndf}{gardennight}{6.29}

\resdef{rendrmse}{ours}{cloister}{8.73}
\resdef{rendrmse}{ours}{matheasy}{7.90}
\resdef{rendrmse}{ours}{mathhard}{8.60}
\resdef{rendrmse}{ours}{quadeasy}{5.85}
\resdef{rendrmse}{ours}{quadmid}{6.27}
\resdef{rendrmse}{ours}{quadhard}{7.91}
\resdef{rendrmse}{ours}{canteenday}{6.16}
\resdef{rendrmse}{ours}{canteennight}{6.24}
\resdef{rendrmse}{ours}{gardenday}{5.53}
\resdef{rendrmse}{ours}{gardennight}{5.51}

\resdef{ate}{fastlio2}{canteenday}{0.070}
\resdef{ate}{fastlio2}{canteennight}{0.078}
\resdef{ate}{fastlio2}{gardenday}{0.073}
\resdef{ate}{fastlio2}{gardennight}{0.080}

\resdef{ate}{sghr}{canteenday}{4.025}
\resdef{ate}{sghr}{canteennight}{3.781}
\resdef{ate}{sghr}{gardenday}{3.663}
\resdef{ate}{sghr}{gardennight}{\resfail}

\resdef{ate}{geotrans}{canteenday}{0.997}
\resdef{ate}{geotrans}{canteennight}{1.676}
\resdef{ate}{geotrans}{gardenday}{0.280}
\resdef{ate}{geotrans}{gardennight}{0.405}

\resdef{ate}{hba}{canteenday}{0.082}
\resdef{ate}{hba}{canteennight}{0.085}
\resdef{ate}{hba}{gardenday}{0.118}
\resdef{ate}{hba}{gardennight}{0.126}

\resdef{ate}{balm}{canteenday}{0.062}
\resdef{ate}{balm}{canteennight}{0.066}
\resdef{ate}{balm}{gardenday}{0.037}
\resdef{ate}{balm}{gardennight}{0.037}

\resdef{ate}{ours}{canteenday}{0.053}
\resdef{ate}{ours}{canteennight}{0.061}
\resdef{ate}{ours}{gardenday}{0.043}
\resdef{ate}{ours}{gardennight}{0.043}

\title{\LARGE \bf
Neural LiDAR Bundle Adjustment
}
\author{Chin Yung Anson Hon$^{1}$, Kaicheng Zhang$^{1}$, and Sen Wang$^{1}$
\thanks{\raggedright $^{1}$The authors are with Imperial College London, London, United Kingdom.
        {\tt\small ch1223@ic.ac.uk; k.zhang23@alumni.imperial.ac.uk; sen.wang@imperial.ac.uk}}%
}
\begin{document}

\maketitle
\thispagestyle{empty}
\pagestyle{empty}

\begin{abstract}
    Recent research has achieved remarkable novel view rendering and scene reconstruction results with Neural Radiance Field (NeRF), including extensions to the LiDAR modality. Few studies have, however, explored the key design differences between RGB NeRFs and LiDAR NeRFs, particularly considering their underlying working principles. In this work, we provide both theoretical and empirical evidence suggesting that the density of volume sampling plays a significant role in LiDAR NeRF. Based on this finding, we propose a novel Neural LiDAR Bundle Adjustment (NeLD-BA) algorithm, which is tailored using efficient volume sampling of LiDAR rays for joint optimization of LiDAR map and poses. Extensive experiments are performed using the Newer College and FusionPortable datasets to demonstrate the proposed NeLD-BA's state-of-the-art performance in multi-view point cloud registration and 3D mapping. Our code is available at \url{https://github.com/SenseRoboticsLab/neld_ba}.
\end{abstract}
\section{Introduction}
Multi-view LiDAR point cloud registration is a fundamental problem in robotics and computer vision, with applications in autonomous driving, SLAM, and 3D reconstruction. Traditional methods for LiDAR point cloud registration typically rely on feature-based or direct methods, which are sensitive to pose initialization, prone to local minima, and dependent on reliable correspondences. These limitations motivate a shift toward differentiable, optimization-based formulations that refine poses directly from raw geometry without relying on explicit correspondences.

Recently, Neural Radiance Fields (NeRFs)~\cite{mildenhall2020nerfOG} have gained significant interest for their powerful capabilities in scene reconstruction and novel view synthesis. A major advancement in RGB image NeRF (RGB NeRF) research is RGB NeRF Bundle Adjustment (BA)~\cite{lin2021barf,wang2022nerfmm,chen2023l2gnerf}, which leverages the differentiability of camera projection to jointly optimize the scene neural representation and camera poses. This has demonstrated remarkable success in recovering accurate camera poses from noisy pose initialization. Nonetheless, few studies have explicitly focused on the problem of LiDAR NeRF-BA, which has the potential to outperform traditional point cloud registration methods.  

Contrary to RGB images which contain rich texture for camera pose optimization, LiDAR point clouds pose the following challenges for LiDAR NeRF-BA: (1) LiDAR point clouds inherently lack textures. A direct implementation of RGB NeRF-BA on LiDAR point clouds would result in sub-optimal pose optimization, as it overlooks the strong geometric cue, i.e., accurate ranges, from LiDAR rays. (2) LiDAR point clouds are spatially sparse. Directly using the volume sampling strategy in RGB NeRFs leaves too few samples near the surface, which weakens the pose gradient and hinders pose optimization, as discussed in Section~\ref{sec:volume_sampling}.

\begin{figure}
  \centering
  \begin{subfigure}[t]{0.49\linewidth}
    \centering
    \includegraphics[width=\linewidth]{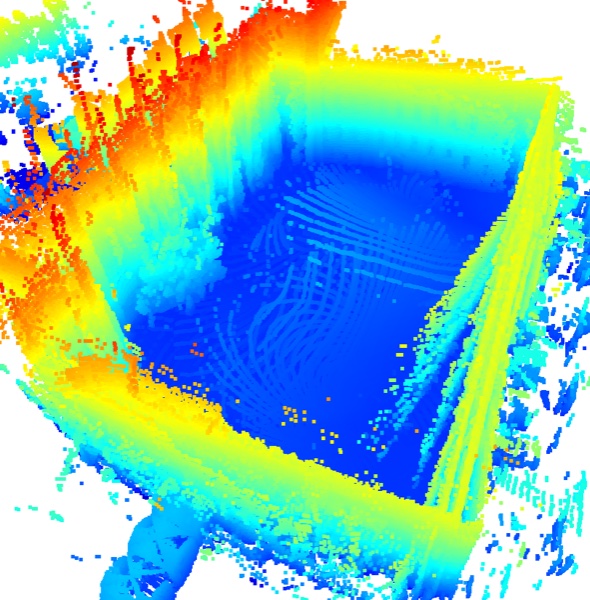}
  \end{subfigure}\hfill
  \begin{subfigure}[t]{0.49\linewidth}
    \centering
    \includegraphics[width=\linewidth]{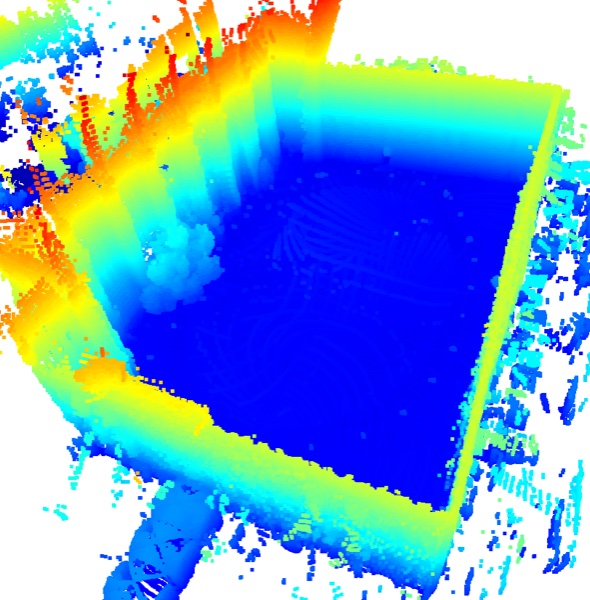}
  \end{subfigure}
  \caption{3D mapping results with HBA~\cite{liu2022HBA} (left) and the proposed NeLD-BA method (right) on the quad\_hard sequence from the Newer College dataset~\cite{Ramezani2020newercollege}. NeLD-BA produces a higher quality map.}
  \label{fig:intro_example}
  \vspace{-10pt}
\end{figure}

To address these challenges, we analyze how volume sampling enters the gradient of the range residual with respect to the sensor pose, and propose \textit{pre-termination fine sampling}, a novel volume sampling strategy tailored to the LiDAR ray model. Given a noisy initialization of LiDAR poses over a sequence, our algorithm, termed NeLD-BA, jointly reconstructs the scene and refines the poses, outperforming existing methods in both novel view synthesis and odometry (Figure~\ref{fig:intro_example}).

Our main contributions in this work include:  

1) We show that volume sampling directly governs the \textit{pose} gradient in LiDAR NeRF-BA, where gradient with respect to the sensor pose is dominated by the few samples straddling the sharp density transition at the surface;  

2) We propose a simple yet efficient strategy for volume sampling and positional encoding tailored to LiDAR NeRF-BA, together with a LiDAR-specific loss function;  

3) Extensive experiments against state-of-the-art algorithms demonstrate superior mapping and localization performance.  

Finally, our code is publicly available on GitHub.

\section{Related Work}
This section reviews both traditional and deep learning methods for point cloud registration. The recent developments of NeRF on LiDAR applications and image/RGB-based NeRF for BA are also reviewed, followed by a focused review on volume sampling for NeRFs.

\subsection{Point Cloud Registration}
Point cloud registration has been one of the long-standing problems in LiDAR research, with applications including LiDAR odometry and 3D mapping~\cite{Zhang2014LOAM}. Traditional techniques for point cloud registration, such as ICP~\cite{besl1992icp} or point to plane ICP~\cite{CHEN1992point2planeicp}, suffer from local minima, while deep learning-based methods~\cite{wang2023sghr,qin2023geotrans,lu2021hregnet}, although achieving state-of-the-art results, suffer from limited generalization and the need for a large amount of point cloud data. Meanwhile, some works study Bundle Adjustment (BA) of LiDAR 3D point clouds~\cite{liu2022HBA,ruhnke2012highly} \cite{liu2021balm}. However, correcting 3D LiDAR points through classic BA methods is challenging due to the inherently unordered and sparse nature of point clouds. 

\subsection{LiDAR NeRF and Neural SDF}
Recent research in NeRF or radiance fields has extended from RGB images to LiDAR data, including novel LiDAR view synthesis~\cite{zheng2024lidar4D,jiang2025gslidar}, mapping and LiDAR odometry~\cite{deng2023nerfloam,isaacson2024lonerlidar}. While these works achieved impressive results, very few have discussed the capability of BA for NeRF. GeoNLF~\cite{xue2024geonlf} has explored the idea of LiDAR NeRF-BA by combining NeRF with a geometry optimizer for LiDAR sensor poses, however, it focused on designing the external architecture and overlooked the underlying mechanism for LiDAR NeRF-BA. LONER~\cite{isaacson2024lonerlidar} proposed a LiDAR NeRF-BA framework that jointly optimizes LiDAR poses and NeRF parameters. However, it did not discuss the influence of volume sampling which closely relates to the gradient behavior of LiDAR NeRF-BA. 

A parallel line of research represents LiDAR scenes with neural signed distance fields (SDFs)~\cite{deng2023nerfloam}\cite{zhong2023shinemapping}\cite{pan2024pinslam}. While neural SDFs produce compact and accurate reconstructions, several of their standard design choices~\cite{takikawa2021neuralsdf} might not be suitable for LiDAR-BA. For example, sphere tracing range rendering could lead to gradient explosion or vanishing during pose optimization due to recursive evaluation on the neural SDF. For these reasons, we build our LiDAR NeRF-BA on density-based volume rendering rather than implicit distance fields.

\subsection{RGB NeRF-BA}
While RGB NeRFs were originally designed for novel-view synthesis with known camera poses~\cite{mildenhall2020nerfOG}, their differentiable nature can also enable gradient-based BA to jointly refine camera poses~\cite{wang2022nerfmm,bian2023nopenerf}. Further research advancements have shown techniques, such as gradual activation of positional encoding frequencies~\cite{lin2021barf} and over-parameterization of camera poses~\cite{chen2023l2gnerf}, can significantly improve BA results. Similar ideas have been applied to image deblurring~\cite{wang2023badnerf}. Although these techniques could inspire LiDAR NeRF-BA, they do not consider the specific characteristics of LiDAR rays.

\subsection{Volume Sampling in NeRF}
Volume rendering of rays requires volume sampling in NeRF~\cite{mildenhall2020nerfOG}, which involves selecting a set of points along the ray to query the NeRF model for volume rendering. In the original NeRF implementation, hierarchical sampling is proposed to first render coarse ray samples before placing fine samples \textit{around} the volumes with high ray termination probability, which we refer to as \textit{centered fine sampling} throughout this paper. This allows for gradient-efficient NeRF model training. Later advancements in research have improved volume sampling to effectively sample conical frusta~\cite{barron2021mipnerf,barron2023zipnerf} or cones on reflected surfaces~\cite{verbin2024nerfcasting}. However, previous research on volume sampling focuses on improving image synthesis quality, with no existing research discussing its significance or design in LiDAR NeRF.

\section{Methodology}\label{sec:methodology}
This section first formulates the LiDAR NeRF-BA problem. We then describe the insights on volume sampling for LiDAR NeRF and the proposed LiDAR NeRF-BA method. 

\subsection{Problem Formulation: LiDAR NeRF-BA}
In general, a NeRF model can be formulated as: 
\begin{equation}\label{eq:NeRF_model}
	\begin{split}
		\hat{\sigma} = f\left(\gamma\left(T(\mathbf{x};\mathbf{p})\right);\Theta\right)
	\end{split}
\end{equation}
where $\hat{\sigma}$ is the estimated density. $T(\cdot)$ denotes the Euclidean transformation function that transforms a volume sample $\mathbf{x} \in \mathbb{R}^3$ by sensor pose $\mathbf{p} \in SE(3)$. $\gamma(\cdot)$ is the positional encoding function, and $f(\cdot)$ is a multi-layer perceptron (MLP) with parameters $\Theta$, representing the NeRF model. 
Note that the viewing direction in the original five-dimensional NeRF input~\cite{mildenhall2020nerfOG} affects only view-dependent radiance, not density. As our model predicts density alone, volume samples remain three-dimensional, although we continue to refer to the model as a NeRF throughout this paper.

For any LiDAR ray denoted as $\mathbf{r}(z)=\mathbf{o}+z\mathbf{u}$, where $\mathbf{o}$ and $\mathbf{u}$ are the ray's origin and direction, with $z \in \mathbb{R}_{>0}$ being the range along the LiDAR ray, an estimated range $\hat{D}$ from the NeRF model can be derived by the alpha compositing function of $N$ volume samples~\cite{mildenhall2020nerfOG}:
\begin{equation}\label{eq:alpha_compositing_part1}
	\hat{D}(\mathbf{r}) = \sum_{i=1}^{N} \hat{h}(z_i) z_i 
\end{equation}
\begin{equation}\label{eq:termination_distribution}
	\hat{h}(z_i) =
	\exp\left(- \sum_{j=1}^{i-1} \hat{\sigma}_j \delta_j \right) 
	\left(1 - \exp\left(-\hat{\sigma}_i \delta_i\right)\right) 
\end{equation}
where $\delta_i$ is the distance between neighboring samples, and $\hat{h}(z)$ is the predicted termination distribution of the ray.

Given $S$ LiDAR scans of point clouds, and the initial estimates of their sensor poses $P = \{\mathbf{p}_s\}_{s=1}^S$, LiDAR NeRF-BA can be formulated as the following optimization problem:

\begin{equation}\label{eq:NeLD_BA_problem_def}
	P^*, \Theta^* = \operatorname*{arg\,min}_{P,\,\Theta}\;
	\mathcal{L}\!\left(\hat{D}(P, \Theta),\; D\right)
\end{equation}
where $D$ is the measured range from LiDAR sensors, and each $\mathbf{p}_s$ is parameterized as the Lie algebra $se(3)$ of $SE(3)$. The optimization goal is to find the set of optimal sensor poses $P^*$ and MLP parameters $\Theta^*$ that minimize a chosen loss function $\mathcal{L}$ (see Section~\ref{sec:loss}). 

\subsection{Volume Sampling for LiDAR NeRF-BA}\label{sec:volume_sampling}

\begin{figure*}[h!]
	\centering
	\includegraphics[width=0.9\textwidth]{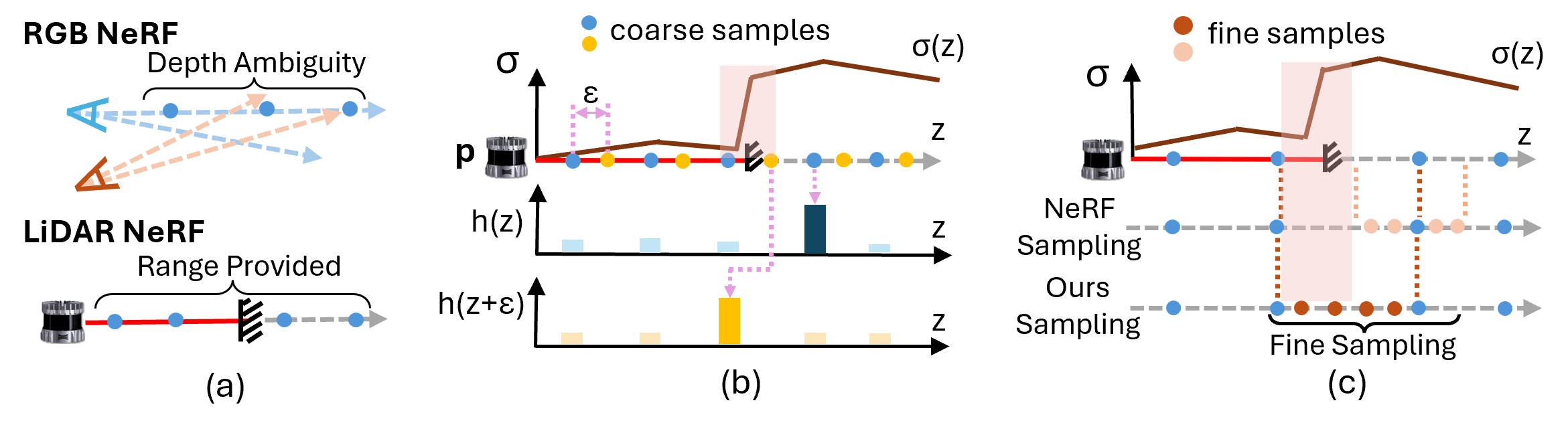}
	\caption{\textbf{Comparison of RGB NeRF-BA and LiDAR NeRF-BA in Volume Sampling along Depth/Range.}
		(a) In RGB NeRF-BA, depth is only \textit{implicitly} constrained via colors and multi-view geometry (orange rays), thus volume samples are loosely constrained. In contrast, LiDAR provides range measurements which require \textit{explicit} volume sampling to constrain. 
		(b) Sampling around the sharp transition of $\sigma(z)$ (pink region) is critical to produce a difference in the ray termination distribution $h(z)$ and $h(z+\epsilon)$ after alpha blending, which provides non-vanishing $\frac{\partial \hat{h}(z_i)}{\partial T(\mathbf{x}_j; \mathbf{p})}$ for pose optimization.
		(c) The centered fine sampling of NeRF~\cite{mildenhall2020nerfOG} allocates fine samples \textit{around} coarse samples, which can miss the sharp transition of $\sigma(z)$ (highlighted pink region). Our pre-termination fine sampling instead places them \textit{prior to} coarse samples, ensuring a dense sampling around the high-probability spike in $\sigma$, akin to a step function.}
	\label{fig:rgb_nerf_ba_vs_lidar_nerf_ba}
\end{figure*}

Compared to RGB NeRF-BA which uses photometric
loss for optimizing camera poses~\cite{lin2021barf,wang2022nerfmm,chen2023l2gnerf}, LiDAR NeRF-BA should make full use of the \textit{range loss} due to the strong geometric cue from LiDAR rays. While RGB NeRF-BA relies on multi-view geometry to optimize poses along depth, LiDAR NeRF-BA should use \textit{range}, as shown in Figure~\ref{fig:rgb_nerf_ba_vs_lidar_nerf_ba} (a).

This motivates us to consider the gradient of the estimated range $\hat{D}$ with respect to the sensor pose $\mathbf{p}$:

\begin{equation}\label{eq:range_gradient}
\frac{\partial \hat{D}}{\partial \mathbf{p}}
=
\sum_{i=1}^{N} z_i
\sum_{j=1}^{i}
\frac{\partial \hat{h}(z_i)}{\partial T(\mathbf{x}_j; \mathbf{p})}
\frac{\partial T(\mathbf{x}_j; \mathbf{p})}{\partial \mathbf{p}}
\end{equation}

We have found that the density of volume sampling, i.e., the number of samples per unit length along the ray, is essential for ensuring the $\frac{\partial \hat{h}(z_i)}{\partial T(\mathbf{x}_j; \mathbf{p})}$ terms do not vanish, as illustrated in Figure~\ref{fig:rgb_nerf_ba_vs_lidar_nerf_ba}(b). We address this through two modifications to volume sampling. 

\begin{figure}
	\centering
	\includegraphics[width=0.47\textwidth]{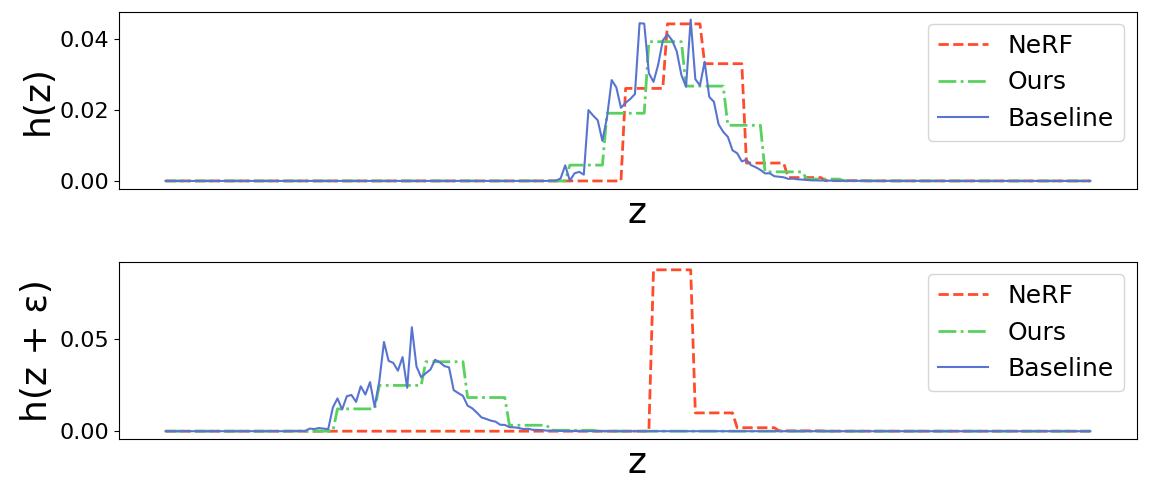}
	\caption{\textbf{Ray Termination Distributions of a LiDAR Ray in canteen\_day sequence} We illustrate the idea in Figure~\ref{fig:rgb_nerf_ba_vs_lidar_nerf_ba}(b) with an actual LiDAR ray. Our pre-termination fine sampling better approximates the baseline distribution constructed with 8192 uniform samples when shifted ($\epsilon=0.01$), which is critical for providing accurate gradients in LiDAR NeRF-BA.}
	\label{fig:ray_termination_dist_comparison}
	\vspace{-15pt}
\end{figure}

First, note that the centered fine sampling strategy of NeRF~\cite{mildenhall2020nerfOG} is sub-optimal for providing gradients for optimizing $\mathbf{p}$. In centered fine sampling, fine samples are placed in a bin \textit{around} coarse samples, guided by weights derived from the coarse termination distribution, as shown in Figure~\ref{fig:rgb_nerf_ba_vs_lidar_nerf_ba}(c). Notably, the sum in Eq.~\eqref{eq:range_gradient} is typically dominated by the first sample after the $\sigma$ spike, and samples after it contribute little to the sum due to alpha blending. Thus, we propose \textit{pre-termination fine sampling}, which places fine samples at volumes \textit{before} coarse samples, as shown in Figure~\ref{fig:rgb_nerf_ba_vs_lidar_nerf_ba}(c). This fully utilizes the fine samples and significantly eliminates the estimated range ambiguity, as shown in Figure~\ref{fig:ray_termination_dist_comparison} with an actual LiDAR ray. 

Second, we followed the standard practice to bound all sampling interval to the normalized cube $C = [-1,1]^3$ to avoid sampling in empty space. We found that this significantly improves the overall volume sampling density and thus the pose optimization results.

\subsection{On Fourier Positional Encoding}
Positional encoding is key to detailed scene reconstruction in both RGB NeRF and LiDAR NeRF~\cite{mildenhall2020nerfOG}. The Fourier positional encoding function $\gamma(\cdot)$ for feature $x$ is~\cite{tancik2020fourierfeatures}:
\begin{align}\label{eq:fourier_positional_encoding_0}
	\gamma(x) & = \left[x, \gamma_0(x), \gamma_1(x), \dots, \gamma_{L-1}(x)\right] \in \mathbb{R}^{1 + 2L}
\end{align}
where $\gamma_k(x) = \left[\cos\left(2^k \pi x\right), \sin\left(2^k \pi x\right)\right] \in \mathbb{R}^{2}$.

There is, however, a $2^k\pi$ amplification in the encoding gradient that makes it difficult to optimize $\mathbf{p}$ and may drive $\mathbf{p}$ away from the global minimum. This is particularly important for the common outdoor use case of LiDAR scenes that uses high-frequency encoding for reconstructing scene details~\cite{martinbrualla2021nerfinwild}. Inspired by the successful use of \textit{surrogate gradient} to optimize spiking neural networks~\cite{Zenke2021surrogate}, the following surrogate gradient for Fourier positional encoding is proposed to handle this problem:

\begin{align}\label{eq:surrogate_fourier_gradient}
	{\frac{\partial \gamma_k^\text{smooth}(x)}{\partial x}}
	= \frac{1}{2^k\pi} {\frac{\partial \gamma_k(x)}{\partial x}}  = \begin{bmatrix}
		- \sin(2^k \pi x) , \cos(2^k \pi x)
	\end{bmatrix}
\end{align}

The normal positional encoding function in Eq.~\eqref{eq:fourier_positional_encoding_0} was used in the forward pass, and the gradient function in Eq.~\eqref{eq:surrogate_fourier_gradient} was used in the backward pass to avoid gradient amplification. Clearly, $\gamma^\text{smooth}_k(\cdot)$ preserves the gradient direction of $\gamma_k(\cdot)$ while removing the amplification component. This allows for using high-frequency Fourier features with no amplification on gradients, ensuring $\mathbf{p}$ stays at the global minimum even for large $L$.
Additionally, we followed BARF~\cite{lin2021barf} in gradually activating Fourier features, and optimization of sensor poses starts only when the $k^*$th frequency for position features is activated. This allows a coarse reconstruction to first be reached by the NeRF model, and the sensor pose can then be more stably optimized towards the global minimum.

\begin{figure*}
	\vspace*{5pt}
	\centering
	\renewcommand{\arraystretch}{0.4}
	\begin{tabular}{@{}c@{\hskip 2pt}c@{\hskip 2pt}c@{\hskip 2pt}c@{\hskip 2pt}c@{\hskip 2pt}c@{}}
		\rotatebox{90}{\parbox[c]{2cm}{\centering\textbf{\small quad\_easy (NC)}}} &
			\includegraphics[width=0.2\textwidth]{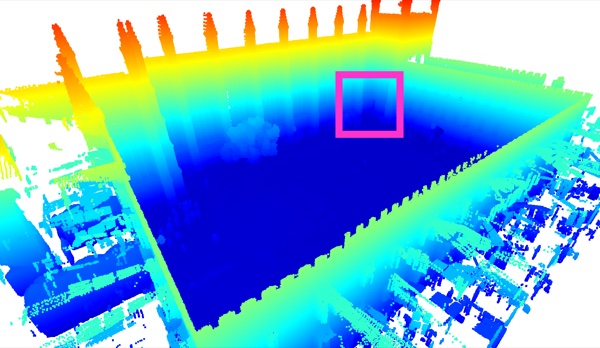} &
			\includegraphics[width=0.18\textwidth]{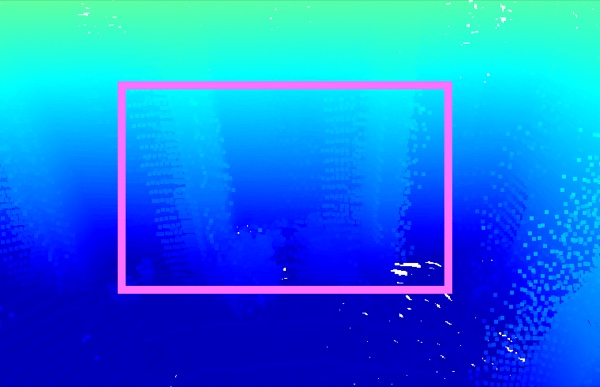} &
			\includegraphics[width=0.18\textwidth]{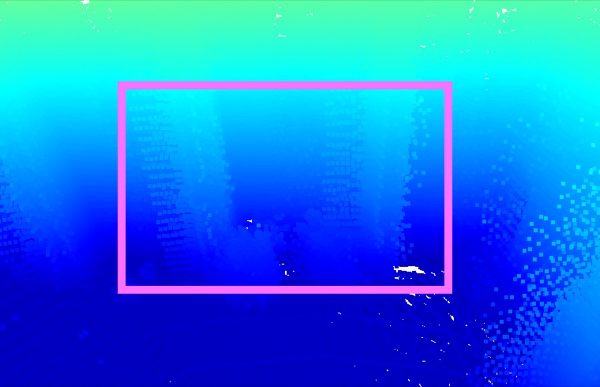}     &
			\includegraphics[width=0.18\textwidth]{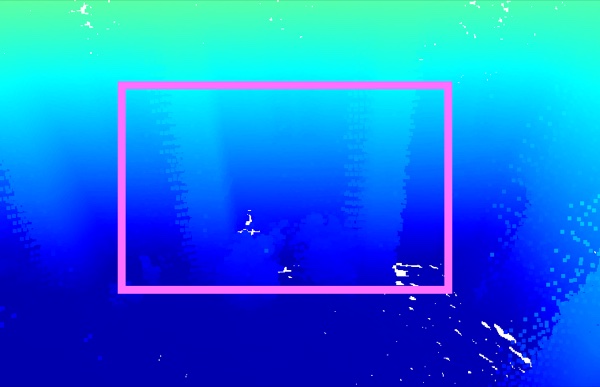}    &
			\includegraphics[width=0.18\textwidth]{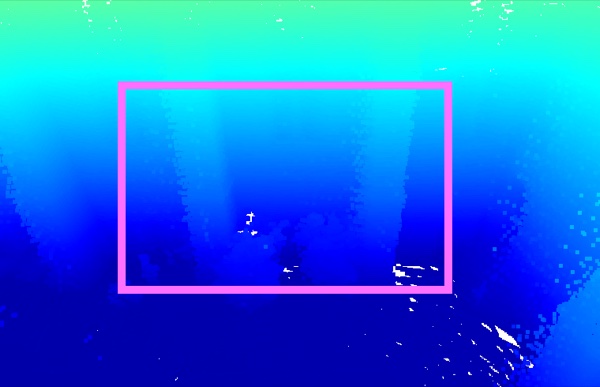} \\

		\rotatebox{90}{\parbox[c]{2cm}{\centering\textbf{\small quad\_hard (NC)}}} &
			\includegraphics[width=0.2\textwidth]{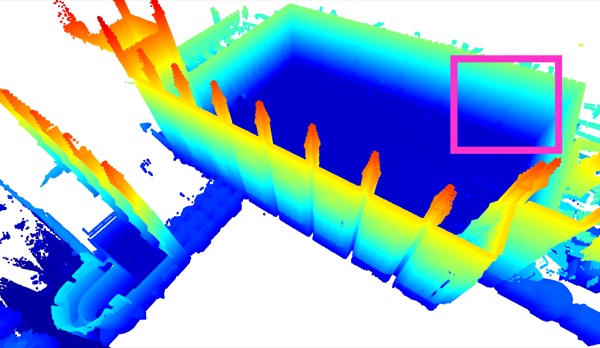} &
			\includegraphics[width=0.18\textwidth]{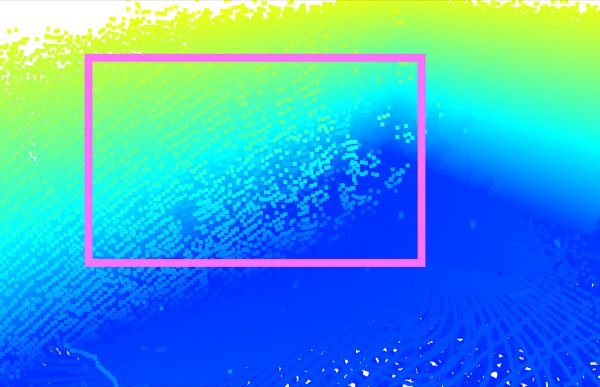} &
			\includegraphics[width=0.18\textwidth]{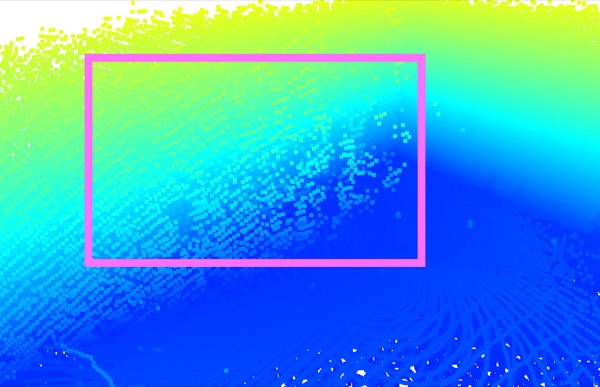}     &
			\includegraphics[width=0.18\textwidth]{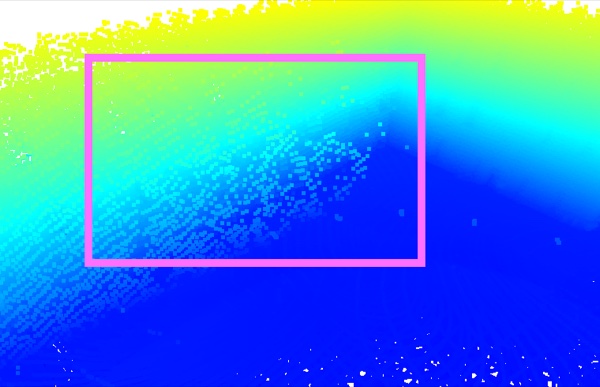}    &
			\includegraphics[width=0.18\textwidth]{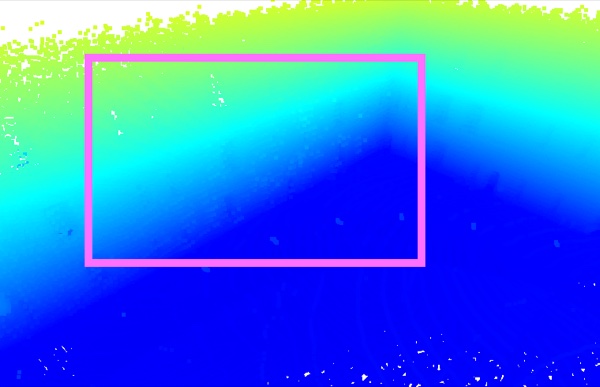} \\

		\rotatebox{90}{\parbox[c]{2cm}{\centering\textbf{\small math\_hard (NC)}}} &
			\includegraphics[width=0.2\textwidth]{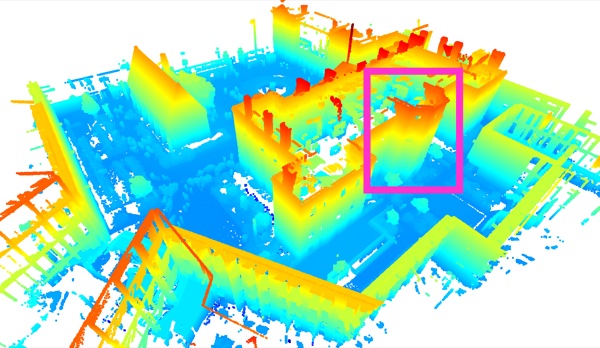} &
			\includegraphics[width=0.18\textwidth]{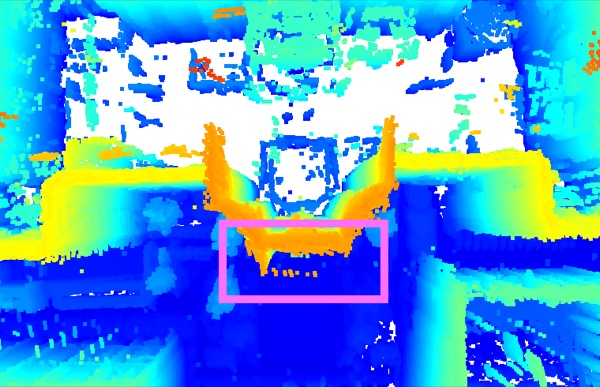} &
			\includegraphics[width=0.18\textwidth]{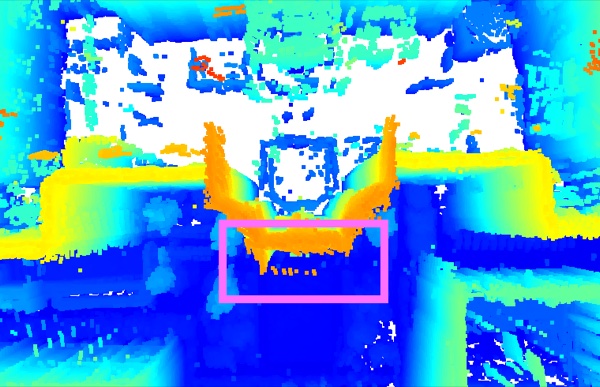}&
			\includegraphics[width=0.18\textwidth]{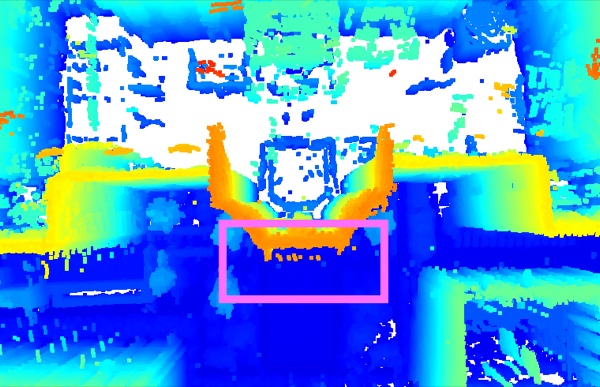}    &
			\includegraphics[width=0.18\textwidth]{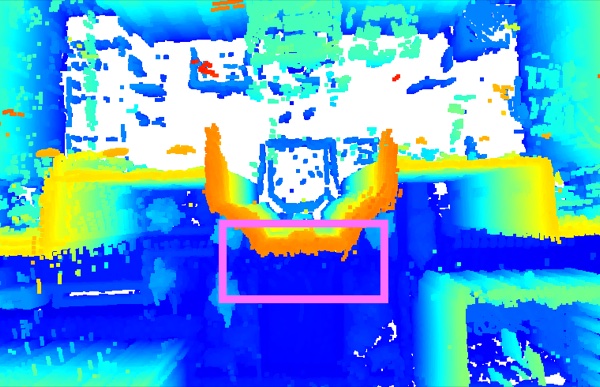} \\

		\rotatebox{90}{\parbox[c]{2cm}{\centering\textbf{\small canteen\_day (FP)}}} &
			\includegraphics[width=0.2\textwidth]{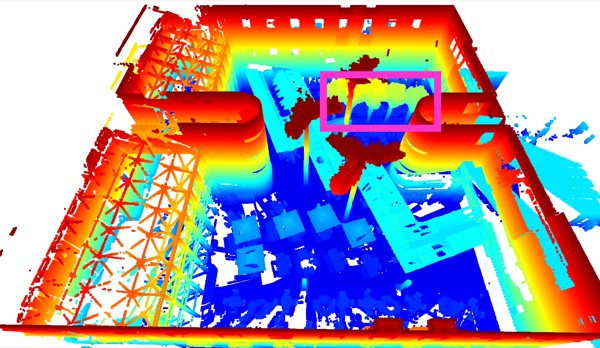} &
			\includegraphics[width=0.18\textwidth]{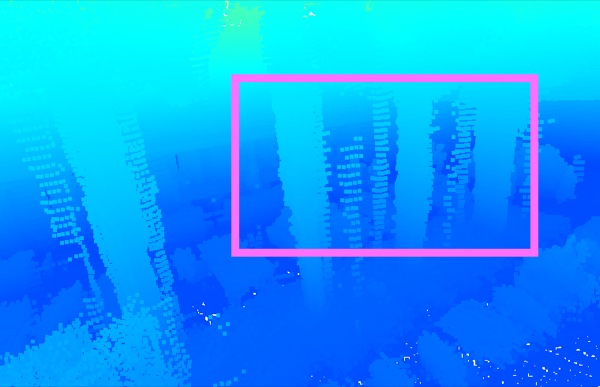} &
			\includegraphics[width=0.18\textwidth]{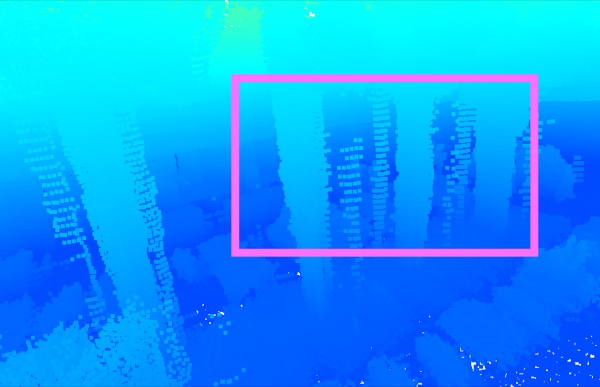}     &
			\includegraphics[width=0.18\textwidth]{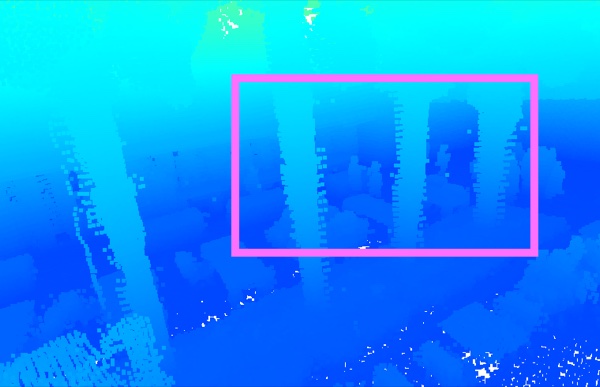}    &
			\includegraphics[width=0.18\textwidth]{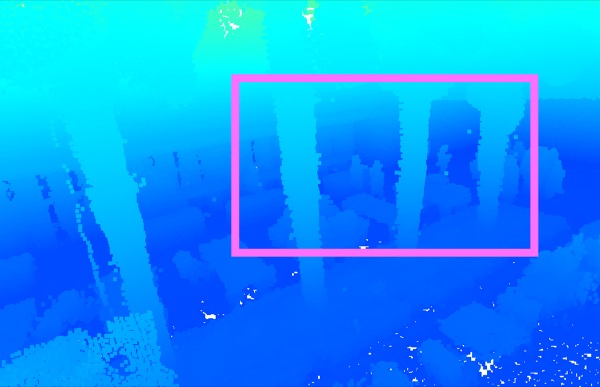} \\

		\rotatebox{90}{\parbox[c]{2cm}{\centering\textbf{\small garden\_day (FP)}}} &
			\includegraphics[width=0.20\textwidth]{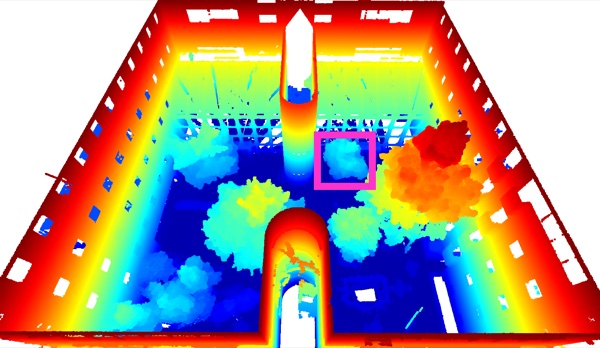} &
			\includegraphics[width=0.18\textwidth]{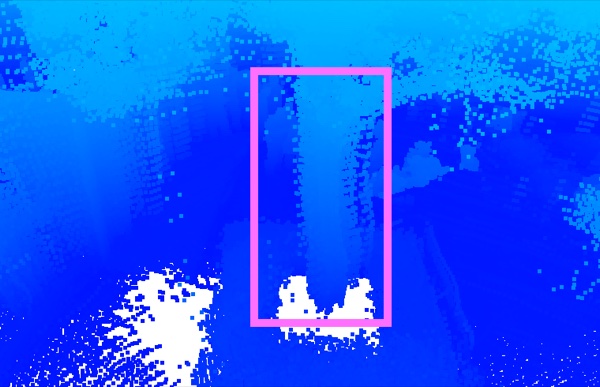} &
			\includegraphics[width=0.18\textwidth]{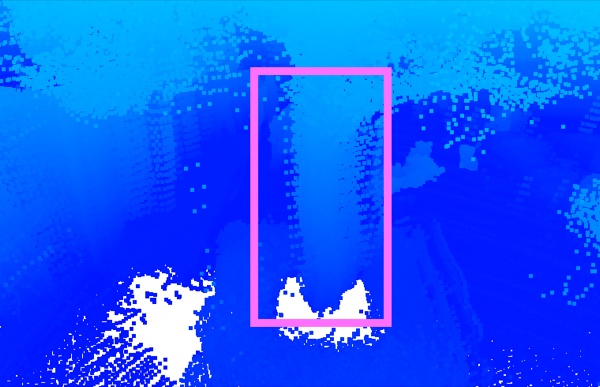}     &
			\includegraphics[width=0.18\textwidth]{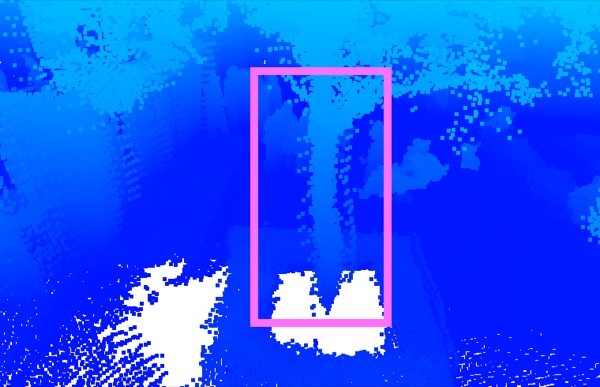}    &
			\includegraphics[width=0.18\textwidth]{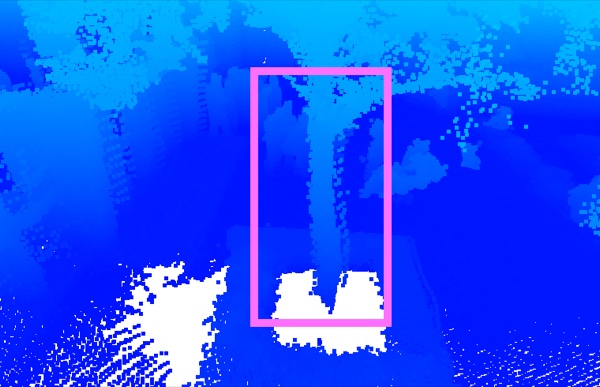} \\[6pt]
		&
        	\small Ground Truth Map &
			\small FAST-LIO2~\cite{xu2021fastlio2} (initial) & 
			\small HBA~\cite{liu2022HBA} & 
			\small BALM ~\cite{liu2021balm} & 
			\small NeLD-BA (ours)\\
	\end{tabular}
	\caption{\textbf{Qualitative Results of Raw Point Cloud Maps for Newer College (NC) and FusionPortable (FP) Sequences.}}
	\label{fig:raw_map_results}
\end{figure*}

\subsection{Loss Functions}\label{sec:loss}
To train the LiDAR NeRF-BA model, the L1 loss was used on both the coarse and fine estimated ranges:
\begin{align}\label{eq:depth_intensity_loss}
	\mathcal{L}_\text{range}  = \sum_\mathbf{r}\left\| \hat{D}(\mathbf{r}) - D(\mathbf{r}) \right\|_1
\end{align}
The distribution loss in DS-NeRF~\cite{deng2024dsnerf} is also used for ray termination distributions, which is the KL divergence loss for comparing the estimated termination distribution $\hat{h}(z)$ against a targeted termination distribution $h^*(z)$:
\begin{align}\label{eq:termination_distribution_loss}
	\mathcal{L}_\text{termination} = \sum_\mathbf{r}\sum_{n=1}^N h^*(z_n) \cdot \ln\left(\frac{h^*(z_n)}{\hat{h}(z_n)}\right)
\end{align}
While DS-NeRF modeled the target termination distribution as a Gaussian distribution with a small variance, we leverage the fact that range is provided in LiDAR measurements and propose a pseudo Dirac delta function as the targeted termination distribution. This effectively enforces that the ray termination probability in the first sample beyond the measured range $D$ is 1:
\begin{align}
	\label{eq:pseudo_dirac_delta}
	h^*(z_n) =
	\begin{cases}
		1, & \text{if } n = \underset{j \in \mathbb{N},\; z_j - D > 0}{\arg\min} { \left\| z_j - D \right\| } \\
		0, & \text{otherwise}
	\end{cases}
\end{align}

Our final loss function $\mathcal{L}$ is:
\begin{align}
\mathcal{L} 
&= \lambda_\text{coarse} \cdot 
   \bigl(\lambda_d \mathcal{L}_\text{range}^c 
       + \lambda_h \mathcal{L}_\text{termination}^c \bigr) \notag \\
&\quad + (\lambda_d \mathcal{L}_\text{range}^f
       + \lambda_h \mathcal{L}_\text{termination}^f)
\label{eq:final_loss}
\end{align}
where superscripts $c$ and $f$ denote loss functions for coarse and fine estimations, and $\lambda$s are hyperparameters to be tuned.

\subsection{Mapping}
A rendered 3D map can be built by casting rays from the optimized sensor poses with the original ray directions in LiDAR frames. Using the same pre-termination fine sampling strategy described above, the termination distances $z_\text{map}$ of rays are determined to be at the position of the highest termination probability per unit length:
\begin{align}
	\label{eq:mapping_with_termination_density}
	i^* = \arg\max_{i} \left( \frac{\hat{h}(z_i)}{\delta_i} \right),
	\qquad
	z_{\text{map}} = z_{i^*}
\end{align}
Finally, we combine points at $\mathbf{r}(z_\text{map})$ from all frames to create a rendered map.

\subsection{Implementation Details}\label{par:implementation_detail}
The same hyperparameters were used for all sequences. In our implementation, each scan was voxelized with a grid size of 0.05 m. 
Our model architecture follows NeRF in the Wild~\cite{martinbrualla2021nerfinwild}, with a hidden layer size of 512, positional encoding of $L=15$ for position features. All models were trained for 160k iterations. Fourier features were gradually activated using BARF's method~\cite{lin2021barf} from iteration 0 to 120k. Sensor poses were unfrozen when the $k^*=6$ frequency for position features was activated.  

The number of coarse and fine samples was increased from 32 to 64 and from 16 to 32 from iteration 0 to iteration 120k. Learning rates were exponentially decayed from $5\times10^{-5}$ to $2\times10^{-5}$. 
The default PyTorch~\cite{paszke2019pytorch} implementation of the Adam optimizer~\cite{kingma2017adam} was used for training. 

Hyperparameters were obtained through grid search with $\lambda_\text{coarse}=0.1$, $\lambda_\text{d}=1$ and $\lambda_\text{h}=1$. Rendered maps were generated using 128 coarse samples and 128 fine samples with our proposed volume sampling strategy. All models took approximately 1.5 hours to train on a single NVIDIA GeForce RTX 4060 GPU.
\begin{table*}
	\caption{Evaluation of Raw Point Cloud Maps Using L1-Chamfer Distance [cm]}
	\label{tab:pcd_plotting_cd-eval}
	\centering
	\renewcommand{\arraystretch}{1.2}
	\begin{tabular}{l@{\hspace{2pt}}*{10}{c@{\hspace{7pt}}}}
		\toprule
		& \multicolumn{6}{c@{\hspace{5pt}}}{Newer College} & \multicolumn{4}{c}{FusionPortable} \\
		\cmidrule(lr){2-7} \cmidrule(lr){8-11}
		Method & 
			\makecell{cloister} &
			\makecell{math\_easy} &
			\makecell{math\_hard} &
			\makecell{quad\_easy} &
			\makecell{quad\_mid} &
			\makecell{quad\_hard} &
			\makecell{canteen\_\\day} &
			\makecell{canteen\_\\night} &
			\makecell{garden\_\\day} &
			\makecell{garden\_\\night} \\
		\midrule
		FAST-LIO2~\cite{xu2021fastlio2} (initial) &
			\underline{\res{rawcd}{fastlio2}{cloister}} &      
			\res{rawcd}{fastlio2}{matheasy} &      
			\res{rawcd}{fastlio2}{mathhard} &      
			\res{rawcd}{fastlio2}{quadeasy} &      
			\res{rawcd}{fastlio2}{quadmid} &      
			\underline{\res{rawcd}{fastlio2}{quadhard}} &      
			\res{rawcd}{fastlio2}{canteenday} &      
			\res{rawcd}{fastlio2}{canteennight} &      
			\res{rawcd}{fastlio2}{gardenday} &      
			\res{rawcd}{fastlio2}{gardennight} \\      
		HBA~\cite{liu2022HBA} &
			\res{rawcd}{hba}{cloister} &      
			\res{rawcd}{hba}{matheasy} &      
			\res{rawcd}{hba}{mathhard} &      
			\res{rawcd}{hba}{quadeasy} &      
			\res{rawcd}{hba}{quadmid} &      
			\res{rawcd}{hba}{quadhard} &      
			\res{rawcd}{hba}{canteenday} &      
			\res{rawcd}{hba}{canteennight} &      
			\res{rawcd}{hba}{gardenday} &      
			\res{rawcd}{hba}{gardennight} \\      
		BALM~\cite{liu2021balm} &
			\textbf{\res{rawcd}{balm}{cloister}} &      
			\textbf{\res{rawcd}{balm}{matheasy}} &      
			\underline{\res{rawcd}{balm}{mathhard}} &      
			\underline{\res{rawcd}{balm}{quadeasy}} &      
			\underline{\res{rawcd}{balm}{quadmid}} &      
			\res{rawcd}{balm}{quadhard} &      
			\underline{\res{rawcd}{balm}{canteenday}} &      
			\underline{\res{rawcd}{balm}{canteennight}} &      
			\underline{\res{rawcd}{balm}{gardenday}} &      
			\underline{\res{rawcd}{balm}{gardennight}} \\      
		NeLD-BA (Ours) &
			\res{rawcd}{ours}{cloister} &      
			\underline{\res{rawcd}{ours}{matheasy}} &      
			\textbf{\res{rawcd}{ours}{mathhard}} &      
			\textbf{\res{rawcd}{ours}{quadeasy}} &      
			\textbf{\res{rawcd}{ours}{quadmid}} &      
			\textbf{\res{rawcd}{ours}{quadhard}} &      
			\textbf{\res{rawcd}{ours}{canteenday}} &      
			\textbf{\res{rawcd}{ours}{canteennight}} &      
			\textbf{\res{rawcd}{ours}{gardenday}} &      
			\textbf{\res{rawcd}{ours}{gardennight}} \\      
		\midrule
		Raw Map w/ GT Pose &
			\res{rawcd}{gtpose}{cloister} &      
			\res{rawcd}{gtpose}{matheasy} &      
			\res{rawcd}{gtpose}{mathhard} &      
			\res{rawcd}{gtpose}{quadeasy} &      
			\res{rawcd}{gtpose}{quadmid} &      
			\res{rawcd}{gtpose}{quadhard} &      
			\res{rawcd}{gtpose}{canteenday} &      
			\res{rawcd}{gtpose}{canteennight} &      
			\res{rawcd}{gtpose}{gardenday} &      
			\res{rawcd}{gtpose}{gardennight} \\      
		\bottomrule
	\end{tabular}
	{\captionsetup{justification=raggedright,singlelinecheck=false}
 \caption*{\small The best results are highlighted in \textbf{bold}, the second-best results are \underline{underlined}.}
 }
 \vspace{-10pt}
\end{table*}

\section{Experimental Results}
In this section, experiments were conducted to compare our method against state-of-the-art algorithms on mapping and trajectory performance. An ablation study is then presented.

\subsection{Experiment Setup}

\begin{figure*}
  \centering
  \renewcommand{\arraystretch}{0.6}
  \begin{tabular}{@{}c@{\hskip 2pt}c@{\hskip 2pt}c@{\hskip 2pt}c@{\hskip 2pt}c@{\hskip 2pt}c@{\hskip 2pt}c@{}}
  &&&&&&      
  \multirow[c]{4}{*}{
        \includegraphics[width=0.069\textwidth]{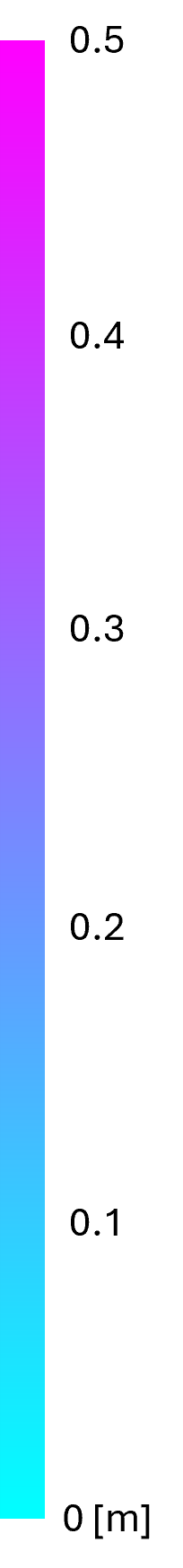}
      }
\\
    \rotatebox{90}{\parbox[c]{2cm}{\centering\textbf{\small quad\_easy (NC)}}} &
      \includegraphics[width=0.175\textwidth]{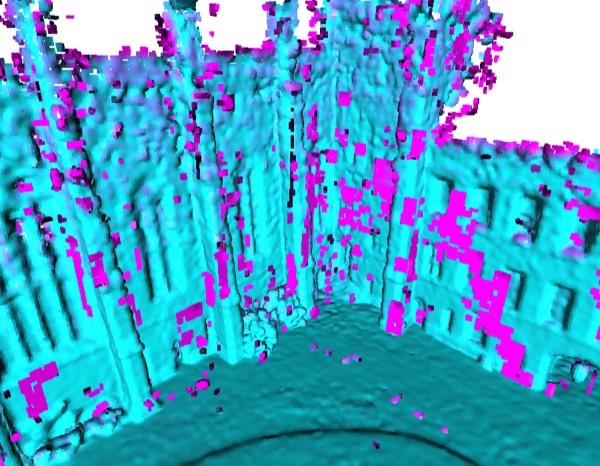} &
      \includegraphics[width=0.175\textwidth]{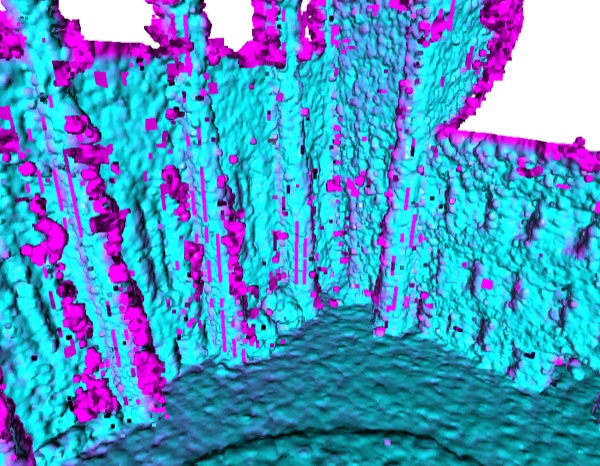} &
      \includegraphics[width=0.175\textwidth]{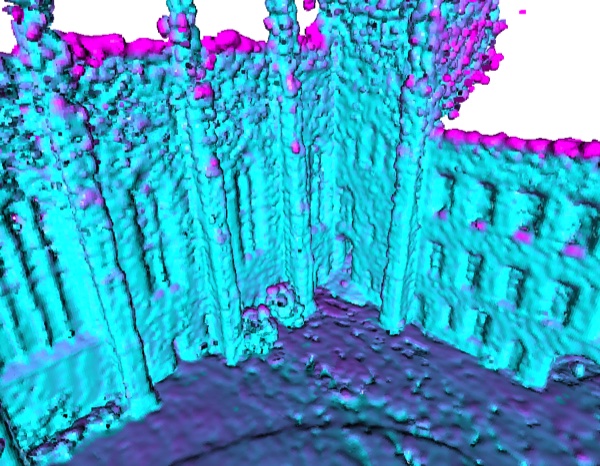} &
      \includegraphics[width=0.175\textwidth]{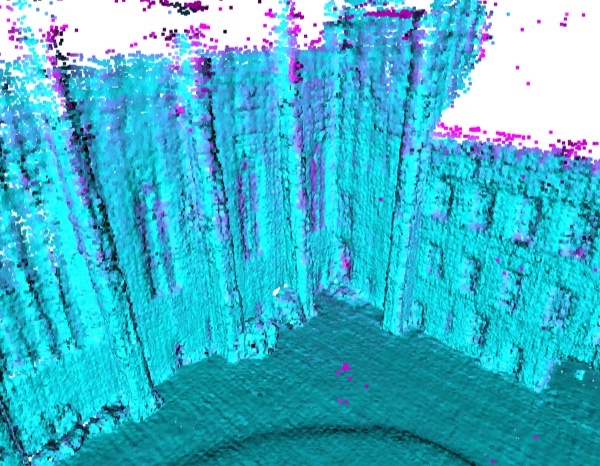} &
      \includegraphics[width=0.175\textwidth]{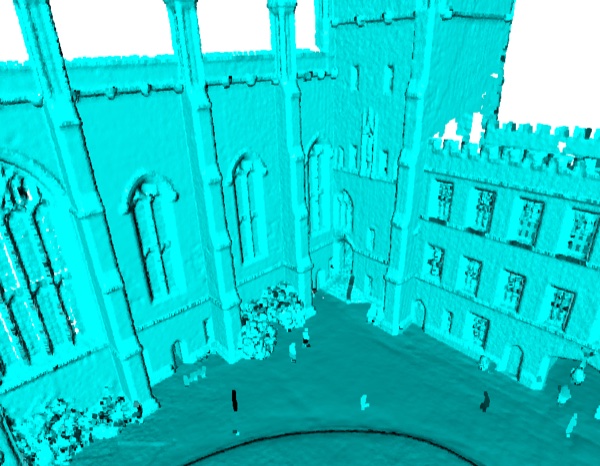} &
 	\\
    \rotatebox{90}{\parbox[c]{2cm}{\centering\textbf{\small quad\_hard (NC)}}} &
      \includegraphics[width=0.175\textwidth]{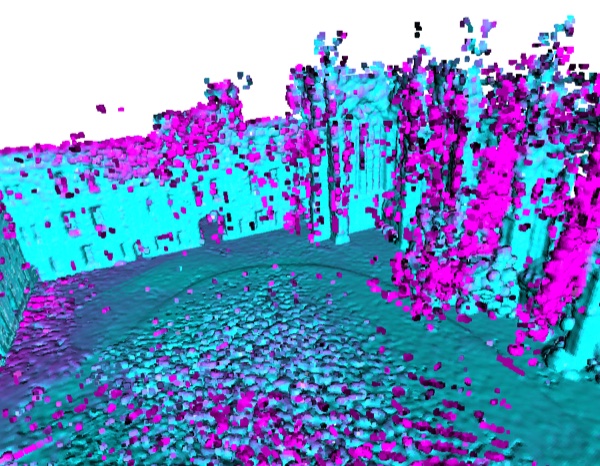} &
      \includegraphics[width=0.175\textwidth]{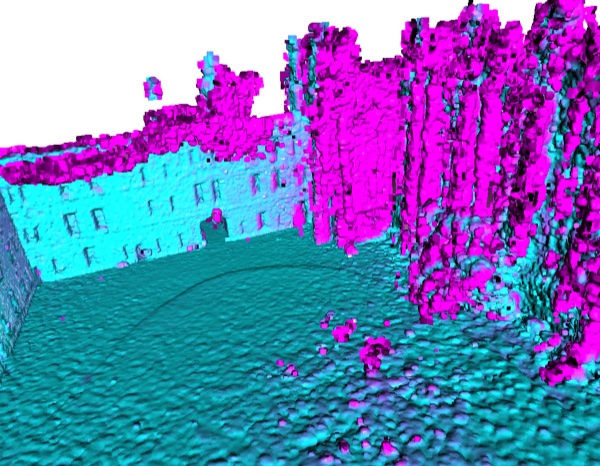} &
      \includegraphics[width=0.175\textwidth]{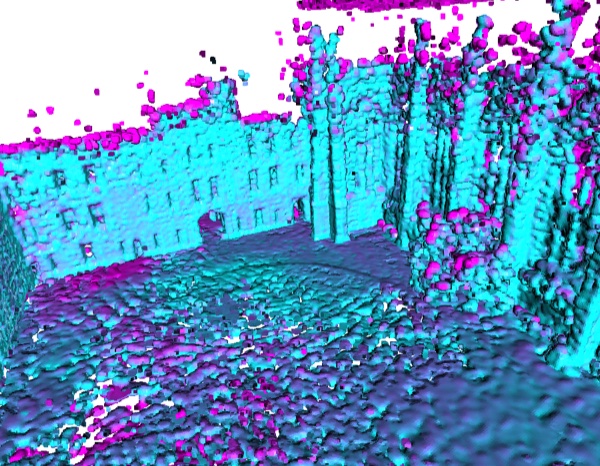} &
      \includegraphics[width=0.175\textwidth]{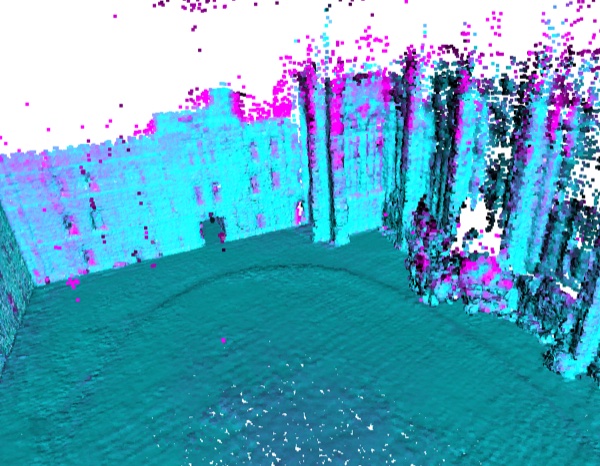} &
      \includegraphics[width=0.175\textwidth]{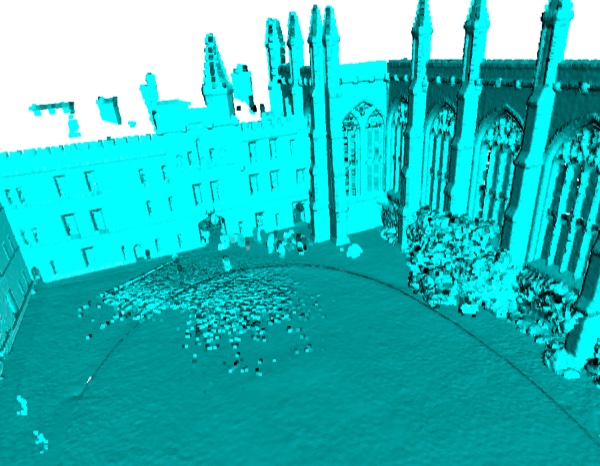} &
      \\ 
    \rotatebox{90}{\parbox[c]{2cm}{\centering\textbf{\small canteen\_night (FP)}}} &
      \includegraphics[width=0.175\textwidth]{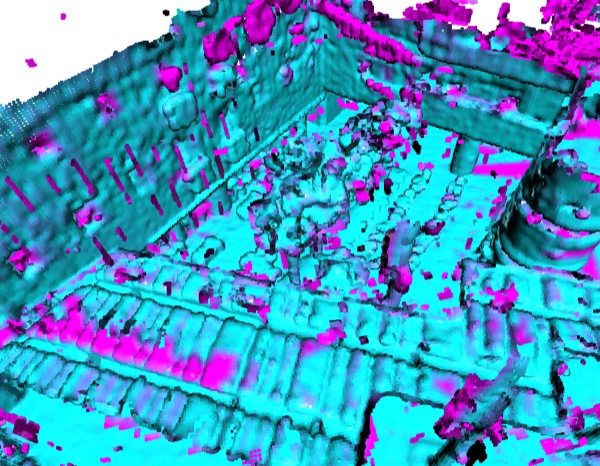}  &
      \includegraphics[width=0.175\textwidth]{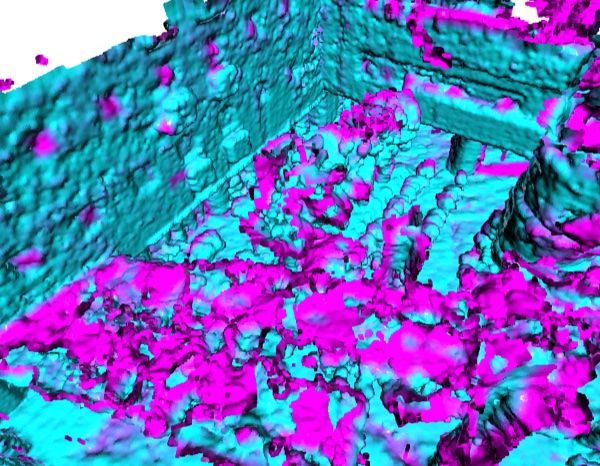}  &
      \includegraphics[width=0.175\textwidth]{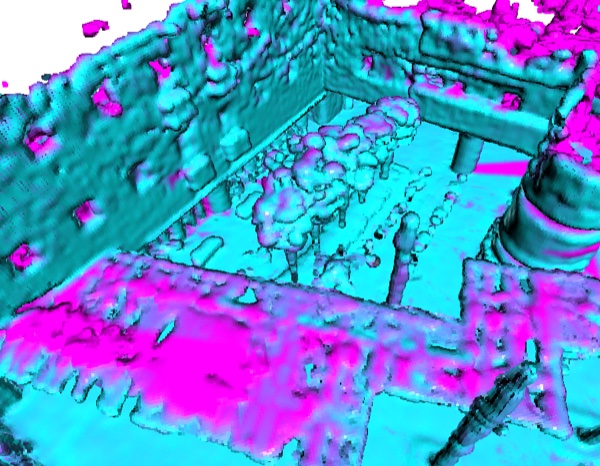}  &
      \includegraphics[width=0.175\textwidth]{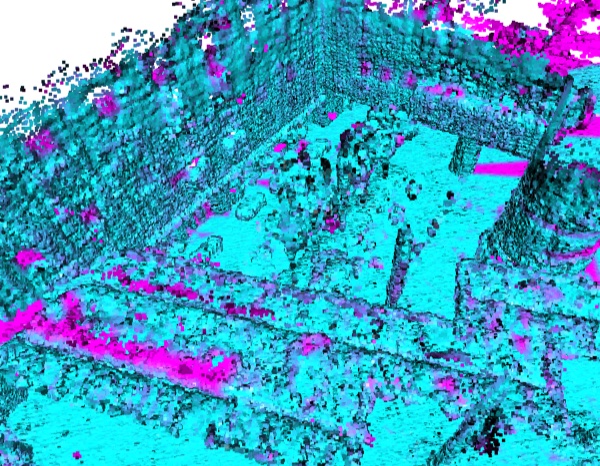}  &
      \includegraphics[width=0.175\textwidth]{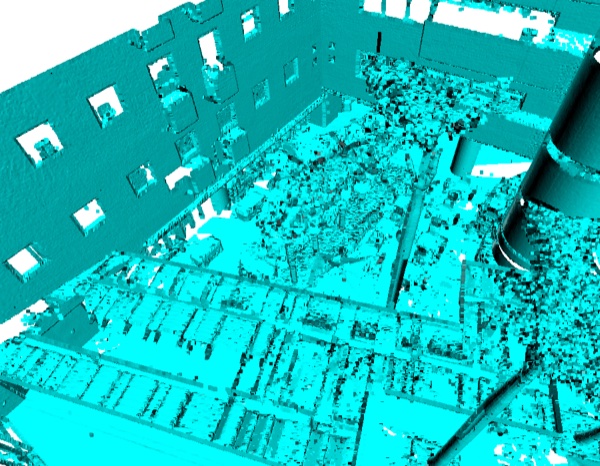}  &
      \\ 
    \rotatebox{90}{\parbox[c]{2cm}{\centering\textbf{\small garden\_night (FP)}}} &
      \includegraphics[width=0.175\textwidth]{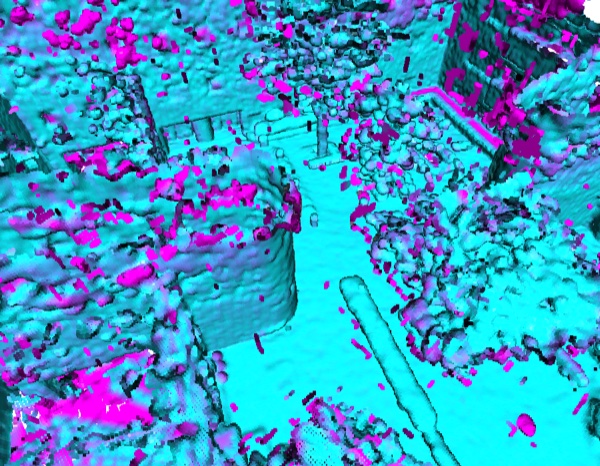} &
      \includegraphics[width=0.175\textwidth]{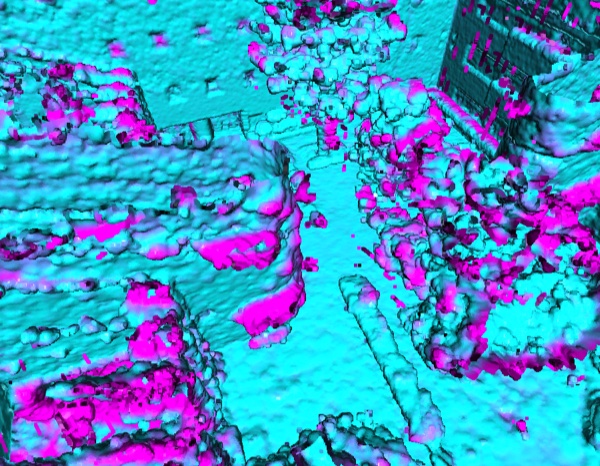} &
      \includegraphics[width=0.175\textwidth]{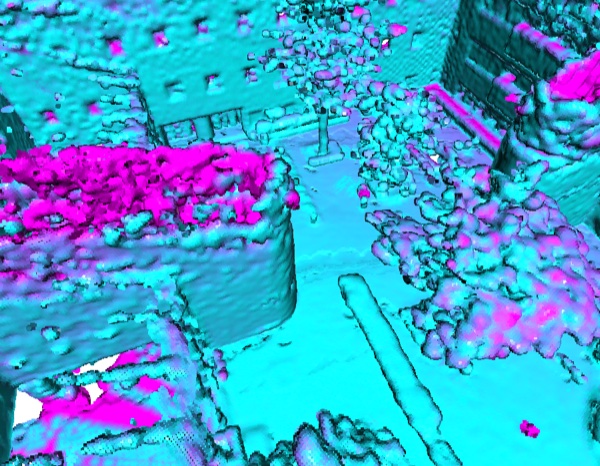} &
      \includegraphics[width=0.175\textwidth]{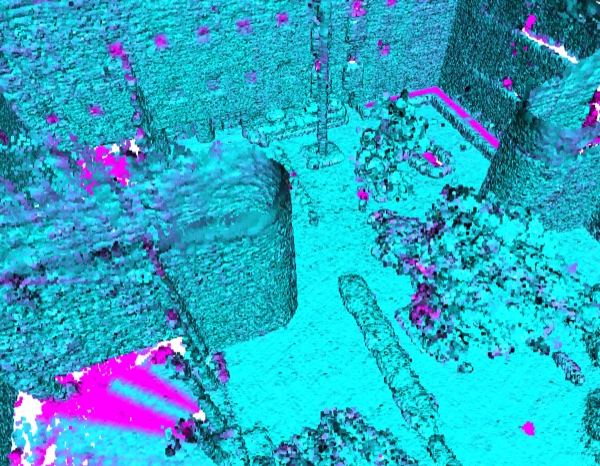} &
      \includegraphics[width=0.175\textwidth]{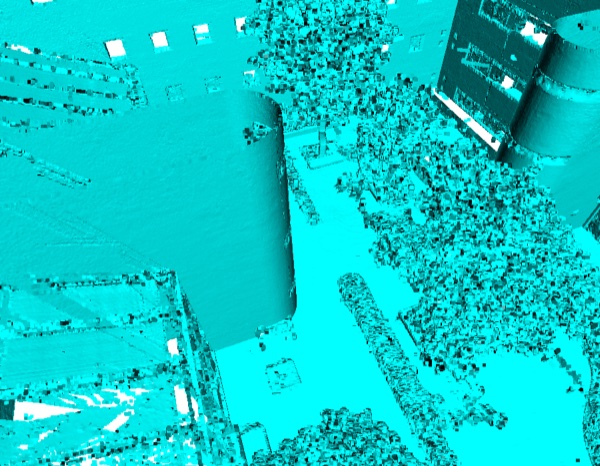} &
      \\ 
    &
      \small SHINE-Mapping~\cite{zhong2023shinemapping} &
      \small PIN-SLAM~\cite{pan2024pinslam} &
      \small 4dNDF~\cite{zhong20244dNDF} &
      \small NeLD-BA (Ours) &
      \small Ground Truth Map &
      \\ 
  \end{tabular}
  \caption{\textbf{Qualitative Results for Rendered Maps of Newer College (NC) and FusionPortable (FP) Sequences.} The color bar on the right shows the L2 distance error of the points with respect to the ground truth map.}
  \label{fig:rendered_map}
\end{figure*}
\begin{figure}
	\centering
	\includegraphics[width=0.48\textwidth]{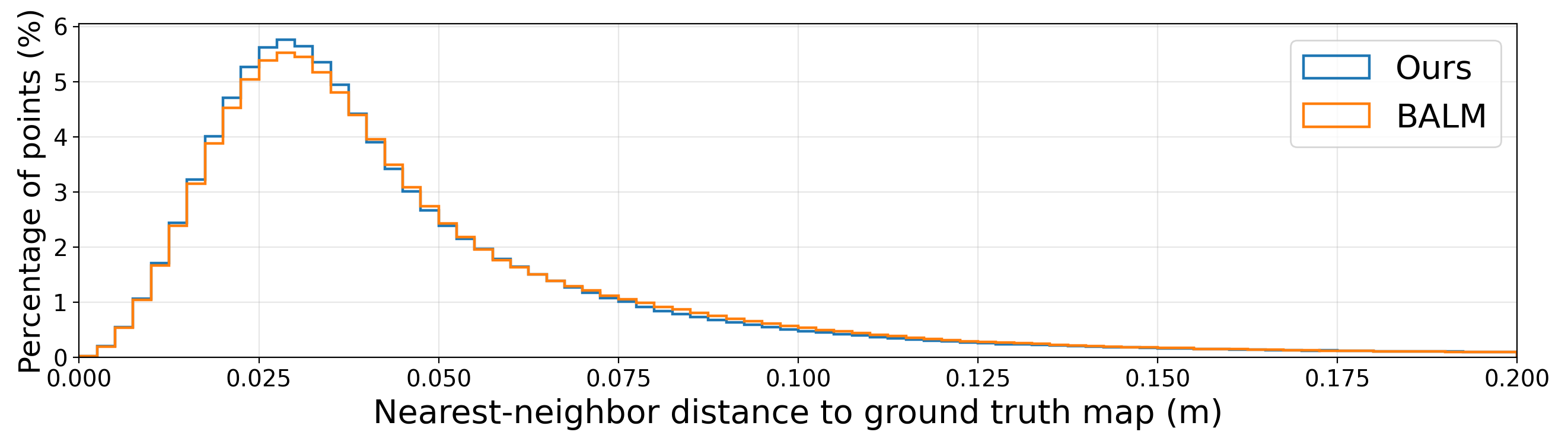}
	\caption{\textbf{Error Histogram on Raw Map.} From the error distribution of garden\_day raw map points, it can be seen that our method results in fewer outlier points than BALM.}
	\label{fig:error_histogram}
	\vspace{-15pt}
\end{figure}

\paragraph{Comparison Algorithms}
We generated raw point cloud maps by registering scans with estimated poses and compared them against the ground truth map to evaluate the poses from HBA~\cite{liu2022HBA}, BALM~\cite{liu2021balm} and our method. The related work GeoNLF~\cite{xue2024geonlf} was not compared due to the lack of support for open-source implementation on the two datasets used. For HBA, BALM and our approach, the same input frames (sampled every 3 seconds) and initial pose estimation from FAST-LIO2 were used. Trajectory evaluation results are provided for HBA, BALM and two deep learning methods, GeoTrans~\cite{qin2023geotrans} and SGHR~\cite{wang2023sghr}. Both GeoTrans and SGHR were evaluated with their official pre-trained models.

We also generated a rendered map from the trained NeRF model, and conducted comparisons against three state-of-the-art LiDAR mapping algorithms with implicit scene representations, including SHINE-Mapping~\cite{zhong2023shinemapping}, PIN-SLAM~\cite{pan2024pinslam}, and 4dNDF~\cite{zhong20244dNDF}. For SHINE-Mapping, we followed the official configuration guidelines to set the frame rate with FAST-LIO2 poses provided as input. For PIN-SLAM, we employed the default configuration for the Newer College dataset, while relying on its internal SLAM to generate poses. For 4dNDF, the same LiDAR scans and FAST-LIO2 poses as provided to our method were used.

\paragraph{Evaluation Metrics}
To quantitatively assess map quality, we employed the L1-Chamfer distance (L1-CD), inlier Root Mean Square Error (RMSE), and the F1 score with a 20 cm threshold when evaluating NeRF-rendered maps against the corresponding ground truth reconstructions. For raw LiDAR-based maps, the L1-Chamfer distance with a 1 m threshold was reported to capture outlier points. All mapping metrics were computed using the FusionPortable~\cite{wei2024fusionportablev2} evaluation framework. For trajectory accuracy, we evaluated the Absolute Trajectory Error (ATE) against the ground truth trajectories. ATE was computed with EVO~\cite{grupp2017evo}, after applying a rigid SE(3) alignment to remove differences due to coordinate frame initialization. This combination of metrics allows us to jointly evaluate both the structural fidelity of the reconstructed environment and the accuracy of the estimated trajectories.

\paragraph{Datasets and Data Processing}
The Newer College~\cite{Ramezani2020newercollege} and FusionPortable~\cite{wei2024fusionportablev2} datasets were used for experiments, since they provide both ground truth (GT) 3D maps from survey-grade LiDAR scanners and ground truth trajectory poses. The FAST-LIO2~\cite{xu2021fastlio2} algorithm was used as an off-the-shelf LiDAR odometry to provide the initial pose estimation, although the proposed method is flexible to work with other algorithms that can estimate LiDAR poses. LiDAR scans and their corresponding FAST-LIO2 poses were sampled every 3 seconds, and only points within a 0.5–80 m sensing range were retained for training.

\subsection{Reconstruction and Mapping Evaluation}\label{sec:pcd_plot_map}
\paragraph{Raw Point Cloud Map Evaluation}
Our method produces a more accurate pose refinement, evident from the substantial reduction in the L1-Chamfer distance of the raw maps from our method compared to those from HBA and BALM, as shown in Table~\ref{tab:pcd_plotting_cd-eval}, particularly on challenging sequences such as quad\_hard and math\_hard. In contrast, HBA often fails to register point clouds effectively, while BALM fails to handle outlier frames, which is also evident in Figure~\ref{fig:error_histogram}. Interestingly, raw maps generated using ground truth poses yield relatively higher L1-Chamfer distances for several Newer College sequences, suggesting that the provided ground truth poses are not perfectly accurate and may introduce additional misalignments as reported in~\cite{zhong20244dNDF}. Qualitative results are shown in Figure~\ref{fig:raw_map_results}. Across sequences with both favorable initializations (e.g., quad\_easy) and challenging ones with poor initial poses (e.g., quad\_hard), our method consistently produces more coherent and structurally faithful raw maps compared to HBA and BALM.

\paragraph{Rendered Map Evaluation}
Benefiting from the pose refinement capabilities, our method is able to produce a more accurate rendered map from NeRF against state-of-the-art mapping algorithms, which is particularly evident in the quad\_hard sequence in Figure~\ref{fig:rendered_map}. Table~\ref{tab:rendered_map} shows the quantitative results for rendered maps. Our algorithm outperforms SHINE-Mapping, PIN-SLAM and 4dNDF on F1 score and L1-CD, while achieving comparable inlier RMSE. For sequences with better initial poses such as quad\_easy, our rendering method is able to produce a cleaner map which reduces outliers, while others over-reconstruct the map, which lowers their F1 score and is also evident in the reduction of high-error points shown in Figure~\ref{fig:rendered_map}.

\begin{table*}
  \caption{Evaluation of Rendered Maps}
  \label{tab:rendered_map}
  \centering
  \renewcommand{\arraystretch}{1.2}
  \begin{tabular}{c l@{\hspace{7pt}}*{10}{c@{\hspace{7pt}}}}
    \toprule
      & & \multicolumn{6}{c@{\hspace{5pt}}}{Newer College} & \multicolumn{4}{c}{FusionPortable} \\
    \cmidrule(lr){3-8} \cmidrule(lr){9-12}
    {Metric} & Method &
      \makecell{cloister} &
      \makecell{math\_easy} &
      \makecell{math\_hard} &
      \makecell{quad\_easy} &
      \makecell{quad\_mid} &
      \makecell{quad\_hard} &
      \makecell{canteen\_ \\ day} &
      \makecell{canteen\_ \\ night} &
      \makecell{garden\_ \\ day} &
      \makecell{garden\_ \\ night} \\

    \midrule
	\multirow{4}{*}{\rotatebox[origin=c]{90}{\shortstack{L1-CD \\ $\downarrow$ [cm]}}} &
    PIN\_SLAM~\cite{pan2024pinslam} &
		\res{rendcd}{pinslam}{cloister} &      
		\res{rendcd}{pinslam}{matheasy} &      
		\res{rendcd}{pinslam}{mathhard} &      
		\res{rendcd}{pinslam}{quadeasy} &      
		\res{rendcd}{pinslam}{quadmid} &      
		\res{rendcd}{pinslam}{quadhard} &      
		\res{rendcd}{pinslam}{canteenday} &      
		\res{rendcd}{pinslam}{canteennight} &      
		\res{rendcd}{pinslam}{gardenday} &      
		\res{rendcd}{pinslam}{gardennight} \\      
    & SHINE-Mapping~\cite{zhong2023shinemapping} &
		\res{rendcd}{shine}{cloister} &      
		\res{rendcd}{shine}{matheasy} &      
		\textbf{\res{rendcd}{shine}{mathhard}} &      
		\underline{\res{rendcd}{shine}{quadeasy}} &      
		\underline{\res{rendcd}{shine}{quadmid}} &      
		\res{rendcd}{shine}{quadhard} &      
		\res{rendcd}{shine}{canteenday} &      
		\res{rendcd}{shine}{canteennight} &      
		\res{rendcd}{shine}{gardenday} &      
		\underline{\res{rendcd}{shine}{gardennight}} \\      
    & 4dNDF~\cite{zhong20244dNDF} &
		\textbf{\res{rendcd}{fourdndf}{cloister}} &      
		\underline{\res{rendcd}{fourdndf}{matheasy}} &      
		\res{rendcd}{fourdndf}{mathhard} &      
		\res{rendcd}{fourdndf}{quadeasy} &      
		\res{rendcd}{fourdndf}{quadmid} &      
		\underline{\res{rendcd}{fourdndf}{quadhard}} &      
		\textbf{\res{rendcd}{fourdndf}{canteenday}} &      
		\textbf{\res{rendcd}{fourdndf}{canteennight}} &      
		\underline{\res{rendcd}{fourdndf}{gardenday}} &      
		\res{rendcd}{fourdndf}{gardennight} \\      
    & NeLD-BA (Ours) &  
		\underline{\res{rendcd}{ours}{cloister}} &      
		\textbf{\res{rendcd}{ours}{matheasy}} &      
		\textbf{\res{rendcd}{ours}{mathhard}} &      
		\textbf{\res{rendcd}{ours}{quadeasy}} &      
		\textbf{\res{rendcd}{ours}{quadmid}} &      
		\textbf{\res{rendcd}{ours}{quadhard}} &      
		\underline{\res{rendcd}{ours}{canteenday}} &      
		\underline{\res{rendcd}{ours}{canteennight}} &      
		\textbf{\res{rendcd}{ours}{gardenday}} &      
		\textbf{\res{rendcd}{ours}{gardennight}} \\      
	\midrule
	\multirow{4}{*}{\rotatebox[origin=c]{90}{F1 $\uparrow$ [\%]}} & 
	PIN\_SLAM~\cite{pan2024pinslam} &
		\res{rendf1}{pinslam}{cloister} &      
		\res{rendf1}{pinslam}{matheasy} &      
		\res{rendf1}{pinslam}{mathhard} &      
		\res{rendf1}{pinslam}{quadeasy} &      
		\res{rendf1}{pinslam}{quadmid} &      
		\res{rendf1}{pinslam}{quadhard} &      
		\res{rendf1}{pinslam}{canteenday} &      
		\res{rendf1}{pinslam}{canteennight} &      
		\res{rendf1}{pinslam}{gardenday} &      
		\res{rendf1}{pinslam}{gardennight} \\      
    & SHINE-Mapping~\cite{zhong2023shinemapping}  &
		\textbf{\res{rendf1}{shine}{cloister}} &      
		\underline{\res{rendf1}{shine}{matheasy}} &      
		\underline{\res{rendf1}{shine}{mathhard}} &      
		\underline{\res{rendf1}{shine}{quadeasy}} &      
		\underline{\res{rendf1}{shine}{quadmid}} &      
		\underline{\res{rendf1}{shine}{quadhard}} &      
		\underline{\res{rendf1}{shine}{canteenday}} &      
		\underline{\res{rendf1}{shine}{canteennight}} &      
		\underline{\res{rendf1}{shine}{gardenday}} &      
		\underline{\res{rendf1}{shine}{gardennight}} \\      
    & 4dNDF~\cite{zhong20244dNDF} &
		\res{rendf1}{fourdndf}{cloister} &      
		\res{rendf1}{fourdndf}{matheasy} &      
		\res{rendf1}{fourdndf}{mathhard} &      
		\res{rendf1}{fourdndf}{quadeasy} &      
		\res{rendf1}{fourdndf}{quadmid} &      
		\res{rendf1}{fourdndf}{quadhard} &      
		\res{rendf1}{fourdndf}{canteenday} &      
		\res{rendf1}{fourdndf}{canteennight} &      
		\res{rendf1}{fourdndf}{gardenday} &      
		\res{rendf1}{fourdndf}{gardennight} \\      
    & NeLD-BA (Ours) & 
		\underline{\res{rendf1}{ours}{cloister}} &      
		\textbf{\res{rendf1}{ours}{matheasy}} &      
		\textbf{\res{rendf1}{ours}{mathhard}} &      
		\textbf{\res{rendf1}{ours}{quadeasy}} &      
		\textbf{\res{rendf1}{ours}{quadmid}} &      
		\textbf{\res{rendf1}{ours}{quadhard}} &      
		\textbf{\res{rendf1}{ours}{canteenday}} &      
		\textbf{\res{rendf1}{ours}{canteennight}} &      
		\textbf{\res{rendf1}{ours}{gardenday}} &      
		\textbf{\res{rendf1}{ours}{gardennight}} \\      
	\midrule
	\multirow{4}{*}{\rotatebox[origin=c]{90}{\shortstack{In. RMSE \\ $\downarrow$ {[cm]}}}} & PIN\_SLAM~\cite{pan2024pinslam} &
		\underline{\res{rendrmse}{pinslam}{cloister}} &      
		\underline{\res{rendrmse}{pinslam}{matheasy}} &      
		\res{rendrmse}{pinslam}{mathhard} &      
		\underline{\res{rendrmse}{pinslam}{quadeasy}} &      
		\textbf{\res{rendrmse}{pinslam}{quadmid}} &      
		\underline{\res{rendrmse}{pinslam}{quadhard}} &      
		\underline{\res{rendrmse}{pinslam}{canteenday}} &      
		\underline{\res{rendrmse}{pinslam}{canteennight}} &      
		\res{rendrmse}{pinslam}{gardenday} &      
		\underline{\res{rendrmse}{pinslam}{gardennight}} \\      
    & SHINE-Mapping~\cite{zhong2023shinemapping}  &
		\res{rendrmse}{shine}{cloister} &      
		\res{rendrmse}{shine}{matheasy} &      
		\textbf{\res{rendrmse}{shine}{mathhard}} &      
		\res{rendrmse}{shine}{quadeasy} &      
		\res{rendrmse}{shine}{quadmid} &      
		\res{rendrmse}{shine}{quadhard} &      
		\res{rendrmse}{shine}{canteenday} &      
		\res{rendrmse}{shine}{canteennight} &      
		\res{rendrmse}{shine}{gardenday} &      
		\res{rendrmse}{shine}{gardennight} \\      
    & 4dNDF~\cite{zhong20244dNDF} &
		\textbf{\res{rendrmse}{fourdndf}{cloister}} &      
		\textbf{\res{rendrmse}{fourdndf}{matheasy}} &      
		\res{rendrmse}{fourdndf}{mathhard} &      
		\res{rendrmse}{fourdndf}{quadeasy} &      
		\res{rendrmse}{fourdndf}{quadmid} &      
		\textbf{\res{rendrmse}{fourdndf}{quadhard}} &      
		\textbf{\res{rendrmse}{fourdndf}{canteenday}} &      
		\textbf{\res{rendrmse}{fourdndf}{canteennight}} &      
		\underline{\res{rendrmse}{fourdndf}{gardenday}} &      
		\res{rendrmse}{fourdndf}{gardennight} \\      
    & NeLD-BA (Ours) & 
		\res{rendrmse}{ours}{cloister} &      
		\res{rendrmse}{ours}{matheasy} &      
		\underline{\res{rendrmse}{ours}{mathhard}} &      
		\textbf{\res{rendrmse}{ours}{quadeasy}} &      
		\underline{\res{rendrmse}{ours}{quadmid}} &      
		\res{rendrmse}{ours}{quadhard} &      
		\res{rendrmse}{ours}{canteenday} &      
		\res{rendrmse}{ours}{canteennight} &      
		\textbf{\res{rendrmse}{ours}{gardenday}} &      
		\textbf{\res{rendrmse}{ours}{gardennight}} \\      
    \bottomrule
  \end{tabular}
  {\captionsetup{justification=raggedright,singlelinecheck=false}
	\caption*{\small The best results are highlighted in \textbf{bold}, the second-best results are \underline{underlined}. --- indicates a failure of the algorithm.}
 }
\end{table*}

\begin{figure*}[!t]
	\centering
	\renewcommand{\arraystretch}{1.0}
	\begin{tabular}{@{}c@{\hskip 2pt}c@{\hskip 2pt}c@{\hskip 2pt}c@{\hskip 2pt}c@{}}&
		\includegraphics[height=0.21\textwidth]{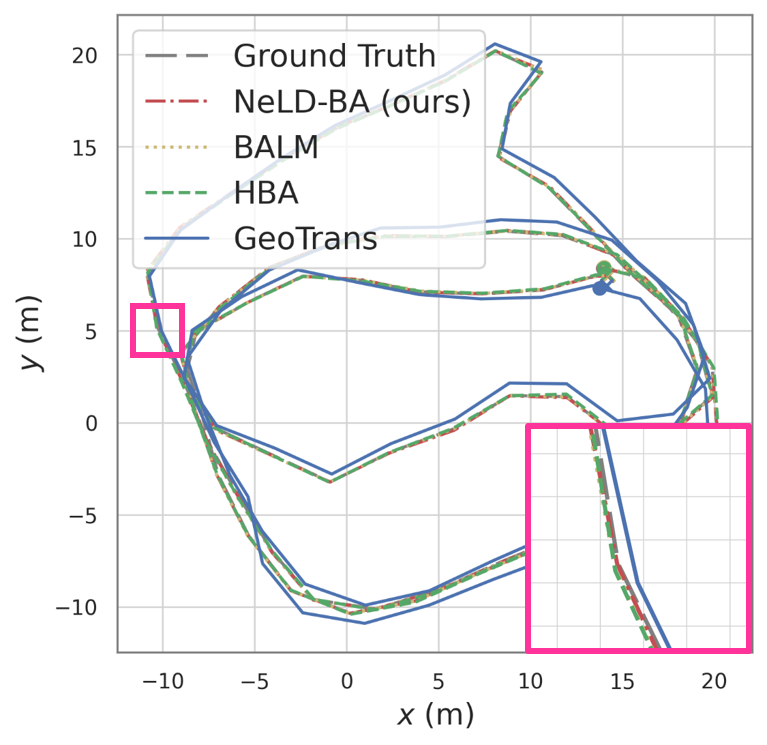}    &
		\includegraphics[height=0.21\textwidth]{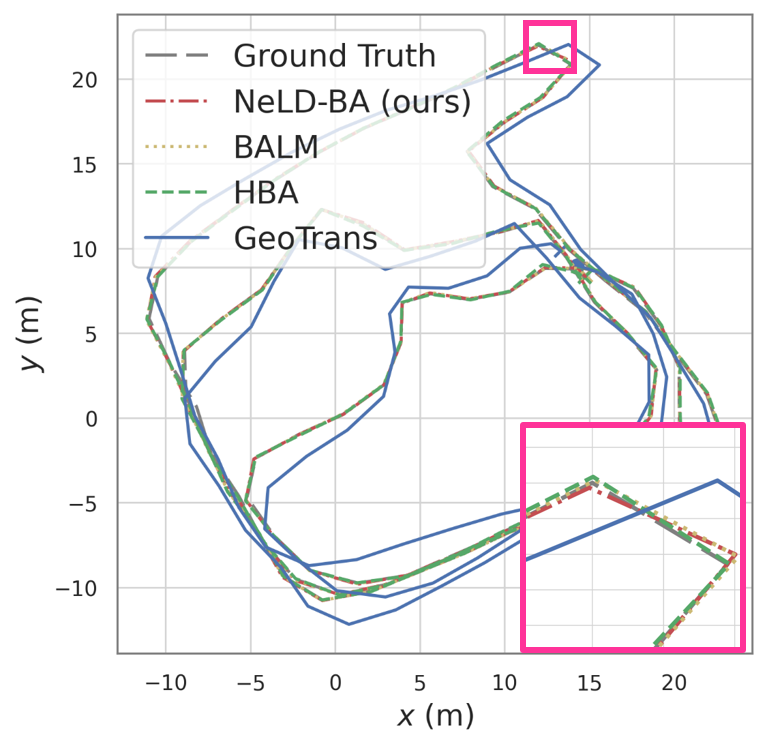} &
		\includegraphics[height=0.21\textwidth]{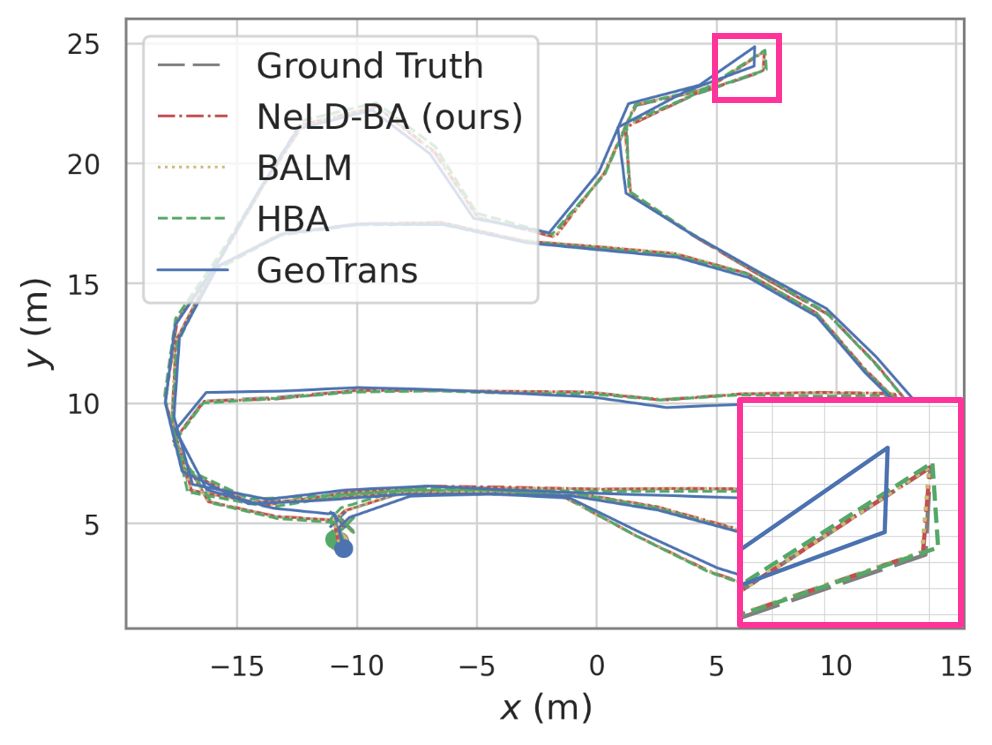}& 
		\includegraphics[height=0.21\textwidth]{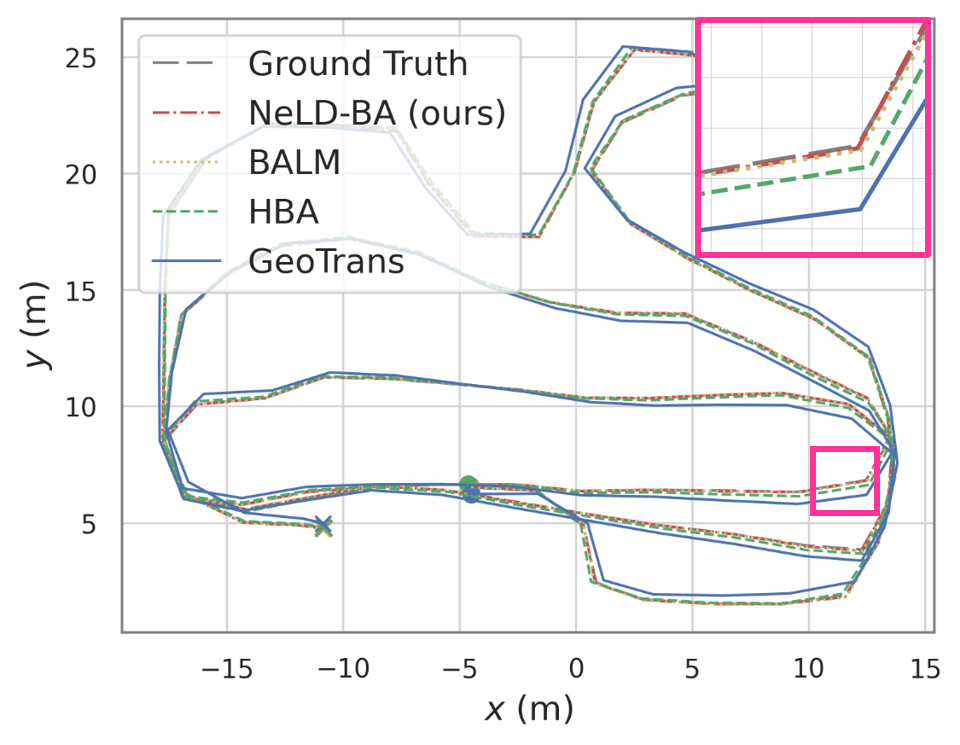}\\
		& \small canteen\_day & \small canteen\_night & \small garden\_day & \small garden\_night\\
	\end{tabular}
	\caption{\textbf{Qualitative Results for Trajectory for FusionPortable Sequences.} Circle markers represent the start of trajectories, cross markers represent the end of trajectories. }
	\label{fig:trajectory_result}
	\vspace{-15pt}
\end{figure*}

\subsection{Trajectory Evaluation}
\begin{table}
	\caption{ATE Mean Values for Trajectory Evaluation for FusionPortable Dataset [meter]}
	\label{tab:ate-eval}
	\centering
	\renewcommand{\arraystretch}{1.2}
	\begin{tabular}{l@{\hspace{2pt}}*{4}{c@{\hspace{5pt}}}}
		\toprule
		Method &
			\makecell{canteen\_\\day} &
			\makecell{canteen\_\\night} &
			\makecell{garden\_\\day} &
			\makecell{garden\_\\night} \\
			\midrule
		FAST-LIO2~\cite{xu2021fastlio2} (initial) &
				\res{ate}{fastlio2}{canteenday} &      
				\res{ate}{fastlio2}{canteennight} &      
				\res{ate}{fastlio2}{gardenday} &      
				\res{ate}{fastlio2}{gardennight} \\      
		SGHR~\cite{wang2023sghr} &
			\res{ate}{sghr}{canteenday} &      
			\res{ate}{sghr}{canteennight} &      
			\res{ate}{sghr}{gardenday} &      
			\res{ate}{sghr}{gardennight} \\      
		GeoTrans~\cite{qin2023geotrans} &
			\res{ate}{geotrans}{canteenday} &      
			\res{ate}{geotrans}{canteennight} &      
			\res{ate}{geotrans}{gardenday} &      
			\res{ate}{geotrans}{gardennight} \\      
		HBA~\cite{liu2022HBA} &
			\res{ate}{hba}{canteenday} &      
			\res{ate}{hba}{canteennight} &      
			\res{ate}{hba}{gardenday} &      
			\res{ate}{hba}{gardennight} \\      
		BALM~\cite{liu2021balm} &
			\underline{\res{ate}{balm}{canteenday}} &      
			\underline{\res{ate}{balm}{canteennight}} &      
			\textbf{\res{ate}{balm}{gardenday}} &      
			\textbf{\res{ate}{balm}{gardennight}} \\      
		NeLD-BA (Ours) &
			\textbf{\res{ate}{ours}{canteenday}} &      
			\textbf{\res{ate}{ours}{canteennight}} &      
			\underline{\res{ate}{ours}{gardenday}} &      
			\underline{\res{ate}{ours}{gardennight}} \\      
		\bottomrule
	\end{tabular}
	{\captionsetup{justification=raggedright,singlelinecheck=false}
 	\caption*{\small The best results are highlighted in \textbf{bold}, the second-best results are \underline{underlined}.}}
	\vspace{-20pt}
\end{table}

We also evaluate the optimized pose with trajectory ATE, and show that our method has substantially improved the initial poses from FAST-LIO2. The quantitative and qualitative results are shown in Table~\ref{tab:ate-eval} and Figure~\ref{fig:trajectory_result}. Both deep learning approaches have underperformed as the pre-trained models struggle to generalize to outdoor scenes. In contrast, BALM and our method have achieved a near ground truth trajectory for FusionPortable sequences. Due to the potentially imperfect ground truth pose for Newer College sequences as discussed in Section~\ref{sec:pcd_plot_map}, trajectory evaluation for Newer College sequences is not provided.

\begin{table}
	\caption{Ablation Study on L1-Chamfer Distance of Raw Point Cloud Maps [cm]}
	\label{tab:ablation_study}
	\centering
	\renewcommand{\arraystretch}{1.2}
	\begin{tabular}{lcccc}
		\toprule 
		& \multicolumn{2}{c@{\hspace{5pt}}}{Newer College} & \multicolumn{1}{c}{FusionPortable} \\
		\cmidrule(lr){2-3}\cmidrule(lr){4-4}
		Method & 
			\makecell{math\_hard} & 
			\makecell{quad\_hard} & 
			\makecell{canteen\_day}\\
		\midrule
		\makecell[l]{w/o Fine Sampling \\ w/o Cube Bound} & 
			\res{rawcd}{ablnohiernocube}{mathhard} &      
			\res{rawcd}{ablnohiernocube}{quadhard} &      
			\res{rawcd}{ablnohiernocube}{canteenday} \\      
		w/o Fine Sampling & 
			\res{rawcd}{ablnohier}{mathhard} &      
			\res{rawcd}{ablnohier}{quadhard} &      
			\res{rawcd}{ablnohier}{canteenday} \\      
		w/o Termination Loss & 
			\res{rawcd}{ablnoterm}{mathhard} &      
			\res{rawcd}{ablnoterm}{quadhard} &      
			\res{rawcd}{ablnoterm}{canteenday} \\      
		w/o Surrogate Grad & 
			\res{rawcd}{ablnosurrogate}{mathhard} &      
			\res{rawcd}{ablnosurrogate}{quadhard} &      
			\res{rawcd}{ablnosurrogate}{canteenday} \\      
		\makecell[l]{Ours w/ Centered Fine \\ Sampling~\cite{mildenhall2020nerfOG}} & 
			\res{rawcd}{ablnerfhier}{mathhard} &      
			\res{rawcd}{ablnerfhier}{quadhard} &      
			\res{rawcd}{ablnerfhier}{canteenday} \\      
		\makecell[l]{Ours w/ Hash Grid} & 
			\res{rawcd}{ablhash}{mathhard} &      
			\res{rawcd}{ablhash}{quadhard} &      
			\res{rawcd}{ablhash}{canteenday} \\      
		NeLD-BA (Ours) & 
			\textbf{\res{rawcd}{ablours}{mathhard}} &      
			\textbf{\res{rawcd}{ablours}{quadhard}} &      
			\textbf{\res{rawcd}{ablours}{canteenday}} \\      
		\bottomrule
	\end{tabular}
	{\captionsetup{justification=raggedright,singlelinecheck=false}
 	\caption*{\small The best results are highlighted in \textbf{bold}.}}
  	\vspace{-20pt}
\end{table}

\begin{figure}
  \centering
  \begin{subfigure}[t]{0.49\linewidth}
    \centering
    \includegraphics[width=\linewidth]{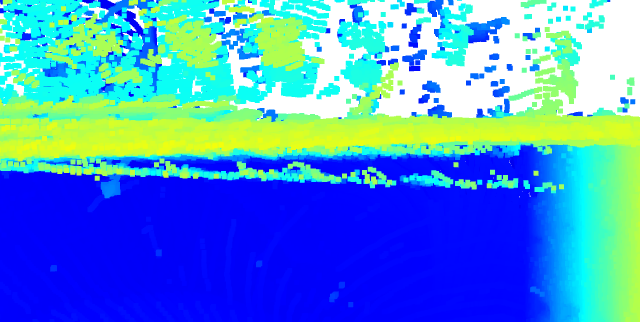}
  \end{subfigure}\hfill
  \begin{subfigure}[t]{0.49\linewidth}
    \centering
    \includegraphics[width=\linewidth]{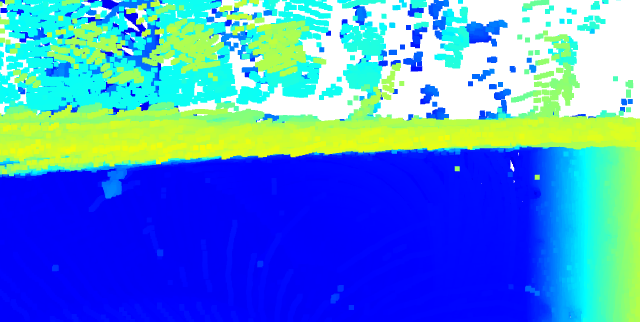}
  \end{subfigure}
  \caption{\textbf{Ablation Study Qualitative Results on Raw Maps.} Centered fine sampling~\cite{mildenhall2020nerfOG} results in an outlier frame (left), while our pre-termination fine sampling successfully registered the outlier frame (right).}
  \label{fig:ablation_qualitative_result}
  \vspace{-20pt}
\end{figure}
\subsection{Ablation Study}
An ablation study was conducted across three sequences to validate our algorithm. We quantitatively compare the ablation results by evaluating the raw point cloud map Chamfer distance, as shown in Table~\ref{tab:ablation_study}. When the volume samples are not bounded within the $[-1,1]^3$ cube and fine sampling is not used, the pose refinement performance significantly drops. It is also evident that the surrogate gradient and termination loss are crucial for achieving high-quality BA. Furthermore, our pre-termination fine sampling is able to eliminate outlier frames in registration compared to NeRF's centered fine sampling, evident in Figure~\ref{fig:ablation_qualitative_result}. Interestingly, using a hashgrid encoding also performed comparable with our method, which is a promising direction for future investigation.

\subsection{Limitations}
Large-scale sequences might be challenging for our method due to the need to normalize point clouds within a $[-1,1]^3$ cube. Furthermore, our algorithm faces a challenge common to NeRFs that it requires relatively long training times.
\section{Conclusions}
This work delves into the mechanics of LiDAR NeRF-BA, showing that volume sampling directly governs the pose gradient, and accordingly proposes pre-termination fine sampling, which densely sample the sharp density transition at the surface to improve pose gradient, together with a surrogate gradient and a termination distribution loss that further enhance BA quality. A promising direction for future work is to explore hash grid features or other representations such as SDFs for LiDAR BA. Finally, our code is publicly available on GitHub to benefit the community.

\bibliographystyle{IEEEtran}
\bibliography{IEEEabrv,references.bib}

\addtolength{\textheight}{-12cm}   







\end{document}